\documentclass[10pt,twocolumn,letterpaper]{article}

\usepackage{cvpr}      

\usepackage{fontawesome}
\usepackage{multirow}
\usepackage{arydshln}
\usepackage{comment}
\usepackage{subcaption}
\usepackage{booktabs}
\definecolor{cvprblue}{rgb}{0.21,0.49,0.74}
\usepackage[pagebackref,breaklinks,colorlinks,allcolors=cvprblue]{hyperref}
\renewcommand{\paragraph}[1]{\noindent\textbf{#1}~}

\usepackage{xcolor}

\definecolor{red}{RGB}{255, 85, 85}  
\definecolor{blue}{RGB}{135, 206, 250} 
\definecolor{green}{RGB}{152, 251, 152}

\title{VQA\textsuperscript{2}: Visual Question Answering for Video Quality Assessment}
\vspace{-15pt}
\author{Ziheng Jia\textsuperscript{1*}$^{\heartsuit}$, Zicheng Zhang\textsuperscript{1*}, Jiaying Qian$^1$,\\ Haoning Wu$^2$, Wei Sun$^1$, Chunyi Li$^1$ Xiaohong Liu$^1$, Weisi Lin$^2$, Guangtao Zhai$^1$,  Xiongkuo Min$^{1\diamondsuit}$ \\
$^1$Shanghai Jiaotong University, $^2$Nanyang Technological University\\
\textbf{\textit{\textcolor{pink}{https://github.com/Q-Future/Visual-Question-Answering-for-Video-Quality-Assessment}}}
}

\begin{document}

\twocolumn[{%
\renewcommand\twocolumn[1][]{#1}%
\maketitle

\begin{center}
    \centering
    \vspace{-2.3em}
    \includegraphics[width=0.93\linewidth]{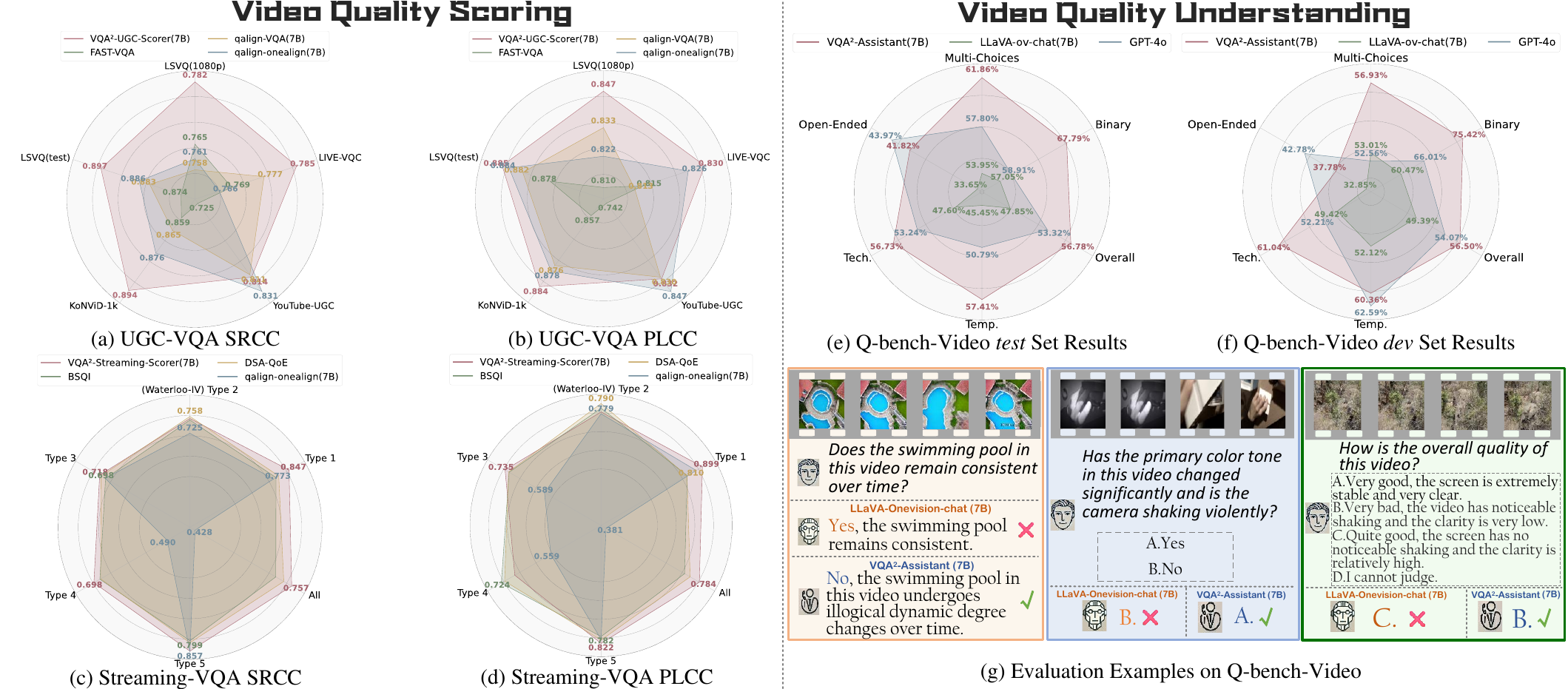}
    \vspace{-10pt}
    \captionof{figure}{The performance of the \textit{\textbf{\textbf{VQA\textsuperscript{2} series models}}} on video quality scoring and video quality understanding tasks. To highlight performance differences, each dimension in all the radar charts is normalized using different ranges based on the specific data characteristics.} \label{fig:performance}
    \vspace{-10pt}
\end{center}%
}]
\let\thefootnote\relax\footnotetext{\noindent\textsuperscript{*}Equal contribution. $^{\heartsuit}$ Project leader. $^\diamondsuit$Corresponding author.}
\begin{abstract}
The advent and proliferation of large multi-modal models (LMMs) have introduced new paradigms to computer vision, transforming various tasks into a unified visual question answering framework. Video Quality Assessment (VQA), a classic field in low-level visual perception, focused initially on quantitative video quality scoring. However, driven by advances in LMMs, it is now progressing toward more holistic visual quality understanding tasks. Recent studies in the image domain have demonstrated that Visual Question Answering (VQA) can markedly enhance low-level visual quality evaluation. Nevertheless, related work has not been explored in the video domain, leaving substantial room for improvement. To address this gap, we introduce the \textbf{\textit{VQA\textsuperscript{2} Instruction Dataset}}—the first \textbf{visual question answering} instruction dataset that focuses on \textbf{video quality assessment}. This dataset consists of  
$3$ subsets and covers various video types, containing $157,755$ instruction question-answer pairs. Then, leveraging this foundation, we present the \textbf{\textit{VQA\textsuperscript{2} series models}}. The \textit{VQA\textsuperscript{2} series models} interleave visual and motion tokens to enhance the perception of spatial-temporal quality details in videos.   We conduct extensive experiments on video quality scoring and understanding tasks, and results demonstrate that the \textit{VQA\textsuperscript{2} series models} achieve excellent performance in both tasks. Notably, our final model, the \textbf{\textit{VQA\textsuperscript{2}-Assistant}}, exceeds the renowned \textit{GPT-4o} in visual quality understanding tasks while maintaining strong competitiveness in quality scoring tasks.
Our work provides a foundation and feasible approach for integrating low-level video quality assessment and understanding with LMMs.
\end{abstract}   

\vspace{-5pt}
\section{Introduction}
\label{sec:intro}
With the invention and rise of large multi-modal models (LMMs) \cite{achiam2023gpt,gao2023llama,zhang2024video,liu2024visual}, the domain of computer vision pertaining to video has entered a new era. Visual Question Answering (VQA) \cite{antol2015vqa}, a pivotal tool for modality alignment, is widely employed in LMM applications. The paradigm of visual-language instruction tuning \cite{liu2024visual} using multi-modal instruction datasets, which encompass vast amounts of high-quality data, has markedly enhanced the performance of video LMMs in high-level visual tasks intimately related to video semantics such as video understanding \cite{zhou2024mlvu,li2024mvbench} and video temporal analysis \cite{li2023seed}. Similarly, in low-level vision, incorporating visual question answering to aid model training and inference has tremendous potential for development. One potential field is Video Quality Assessment (VQA), which is closely related to video quality attributes (such as flicker, blur, and stuttering). We believe integrating visual question answering into developing video quality assessment models can deliver superior quantitative assessment and quality understanding capabilities compared to conventional models, thus holding greater potential for broad applications. The model can be utilized in video encoding, transmission, and decoding processes \cite{kua2017survey}, providing effective feedbacks. Furthermore, it holds promise in the image/video generation domain as effective guidance for refining local generation details \cite{li2024g}.

Concurrently, most video quality assessment models focus solely on video quality scoring \cite{min2024perceptual}; however, they entirely lack the capability to understand and analyze quality details, resulting in significant deficiencies in model versatility. Moreover, existing models with low-level visual quality understanding function almost exclusively apply to the image field \cite{zhang2024quality}, lacking effective perception proficiencies for video-specific temporal and motion quality attributes. To bridge this gap, we construct the \textit{\textbf{VQA\textsuperscript{2} Instruction Dataset}}—a large-scale instruction dataset specifically for video quality assessment based on visual question answering. The dataset lays a solid foundation for developing robust video quality assessment models with remarkable versatility.  The construction pipeline can be divided into $3$ stages along with their corresponding subsets:
\begin{itemize}
\item \textit{Stage-1: Subset centered on distortion recognition for model pre-training.} We develop a distortion recognition instruction subset for model pre-training, leveraging the distortion information from multiple existing datasets.
\item \textit{Stage-2: Instruction tuning subset centered on video quality scoring.} We utilize the mean opinion scores (MOSs) from various existing datasets and transform them into quality-level labels serving as instruction data.
\item \textit{Stage-3: Instruction tuning subset for video quality understanding.} We curate a high-quality, diverse dataset expanded by \textit{GPT} following human expert annotations.

\end{itemize}

Our core contributions are as follows:
\textbf{1)} We construct the \textit{first} \textbf{visual question answering} based instruction-tuning dataset for \textbf{video quality assessment} --- the \textit{\textbf{VQA\textsuperscript{2} Instruction Dataset}}. The dataset encompasses $3$ subsets and includes over $150,000$ instruction pairs, covering various video types such as user generated content (UGC), streaming, and artificial intelligence generated content (AIGC) videos. This ensures both data adequacy and diversity.
 \textbf{2)} We design a complete training strategy and introduce the \textit{\textbf{VQA\textsuperscript{2} series models}}, including the \textbf{\textit{VQA\textsuperscript{2}-Scorers}} and the  \textbf{\textit{VQA\textsuperscript{2}-Assistant}}. The \textit{VQA\textsuperscript{2}-Scorers} achieve state-of-the-art (SOTA) performance in multiple video quality scoring tasks. Meanwhile, the \textit{VQA\textsuperscript{2}-Assistant} excels in video quality understanding tasks, outperforming the proprietary \textit{GPT-4o} on relevant benchmark tests. It also maintains robust performance in quality scoring tasks, showcasing the model's functional versatility and adaptability. The performance and some function examples of our \textit{VQA\textsuperscript{2} series models} are summarized in Fig. \ref{fig:performance}.

%
\vspace{1pt}
\section{Related Works}


\subsection{Video Quality Assessment}
Classical video quality assessment tasks heavily rely on the MOSs obtained from subjective experiments. Two significant tasks among them are the UGC video and streaming video quality scoring tasks. UGC video quality scoring task involves datasets like \cite{nuutinen2016cvd2014,hosu2017konstanz,sinno2018large,wang2019youtube,ghadiyaram2017capture,ying2021patch}, which contain various authentic or synthetic distortions. Streaming video quality scoring datasets include the Waterloo-SQoE series \cite{duanmu2016quality,duanmu2017quality,duanmu2018quality,duanmu2020waterloo} and the LIVE-NFLX series \cite{bampis2021towards,bampis2017study,ghadiyaram2017subjective}, which mainly use simulated transmission distortions (such as rebuffering, long-term stalling, and bitrate switching).




Classic video quality assessment models can be broadly divided into knowledge-driven and data-driven approaches. Knowledge-driven models \cite{mittal2015completely,mittal2012making,saad2014blind,tu2021ugc,tu2021rapique,duanmu2016quality,duanmu2023bayesian} rely on elaborately designed features to evaluate video quality. In contrast, data-driven models \cite{li2022blindly,li2019quality,liu2023ada,liu2018end,sun2022deep,sun2024analysis,wang2021rich,wu2023neighbourhood,ying2021patch,zhang2023md,wen2024modular,bampis2018feature,jia2024continuous} primarily employ deep neural networks (DNNs) to extract features sensitive to video quality. With the widespread application of LMMs in low-level vision, recent works based on LMMs have achieved higher performance. For example, q-align \cite{wu2023q} attains high precision and generalizability in video quality scoring. 

However, these models possess only the capability to score the video quality but almost entirely miss the function of video quality understanding and analysis, with no capability to provide reasonable responses to diverse question types. Thus, they can not realize the boosting demand for quality understanding and analysis of spatial and temporal quality attributes in videos.  Our proposed \textit{VQA\textsuperscript{2}-Assistant} can perform precise video quality scoring while exhibiting strong capabilities in video quality understanding and question answering, marking new progress in this field.

\subsection{Low-level Visual Question Answering}

In the field of low-level visual question answering for images, recent research has made significant advances. Q-Bench \cite{zhang2024q1} is an image-centered visual question-answering-based benchmark for evaluating LMMs on low-level quality understanding tasks. Q-Instruct \cite{wu2024q} has substantially enhanced LMMs' capabilities in understanding the low-level visual quality of images by constructing a large and diverse instruction dataset through human annotation. DepictQA \cite{you2023depicting} leverages LMMs to provide detailed, language-driven evaluations that outperform traditional score-based methods. AesExpert \cite{huang2024aesexpert} builds expert-level aesthetic foundational models by assembling a rich corpus of image aesthetic critique datasets. Co-Instruct \cite{wu2025towards} has significantly improved large models' ability to analyze the quality of multiple visual stimuli by creating instruction datasets based on multi-image comparisons and joint analysis.

However, for videos, no existing work has incorporated low-level visual question answering to create models with enhanced video quality understanding ability. Our work is the first to achieve this, thus paving a promising way for deep video quality analysis through LMMs.

\label{sec:formatting}

\begin{figure}[t]
  \centering
  \includegraphics[width=0.83\linewidth]{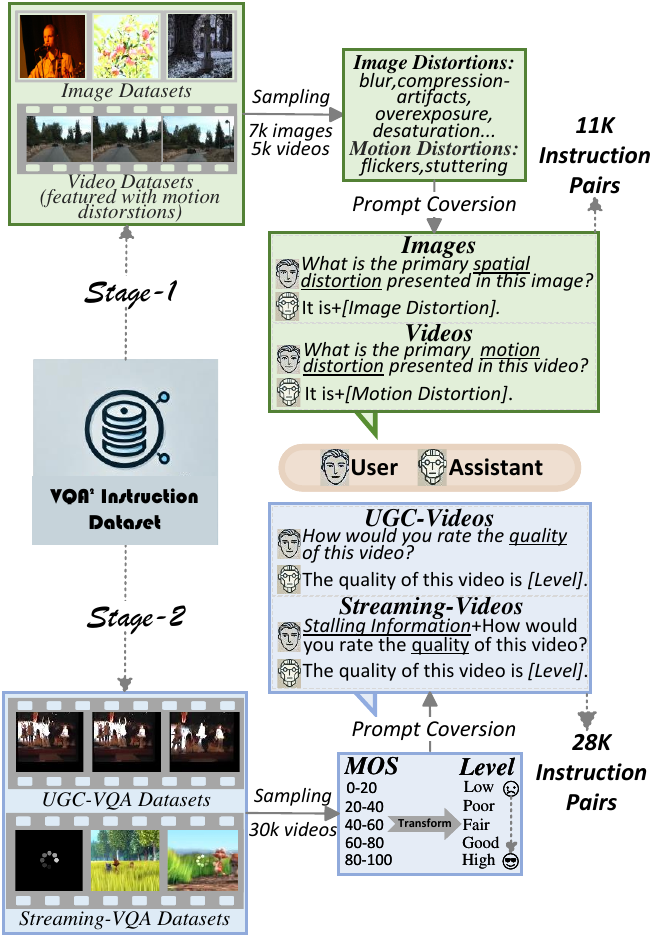}
    \vspace{-5pt}
   \caption{Data construction pipelines of \textit{Stage-1} and \textit{Stage-2}. }
     \vspace{-15pt}
   \label{fig:stage12}

\end{figure}
\begin{figure*}[h]
  \centering
  \includegraphics[width=0.95\linewidth]{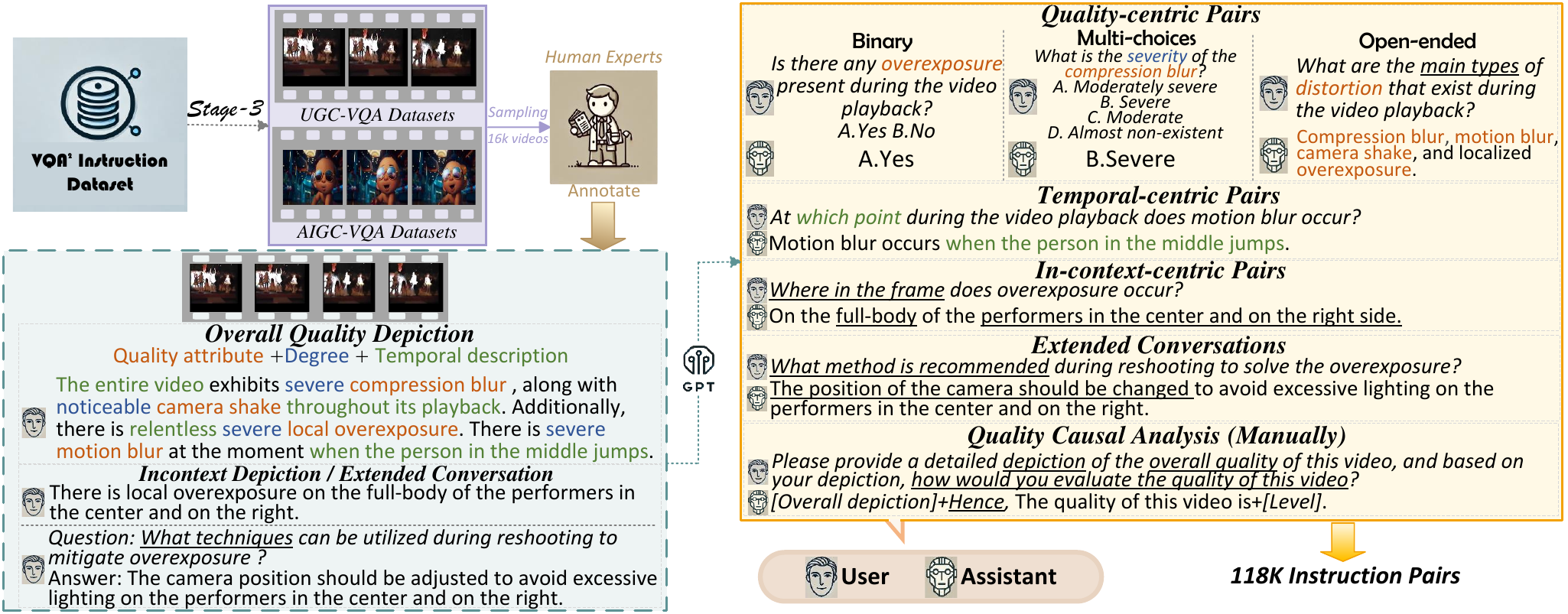}
    \vspace{-5pt}
   \caption{Data construction pipeline of \textit{Stage-3}. This subset is annotated by human experts and then refined and expanded through \textit{GPT}.}
   \vspace{-10pt}
   \label{fig:stage3}
\end{figure*}

\section{The VQA\textsuperscript{2} Instruction Dataset}
As the foundation of our work, we propose the \textit{VQA\textsuperscript{2} Instruction Dataset}.
The dataset construction is composed of $3$ stages: \textit{Stage-1} is used for distortion recognition pretraining, while \textit{Stage-2} and \textit{Stage-3} focus on improving the model's capabilities in video quality scoring and video quality understanding, respectively. The data construction pipelines are shown in Fig. \ref{fig:stage12} and Fig. \ref{fig:stage3}.
\vspace{-2pt}
\subsection{Video Selection or Sampling}
To ensure the diversity of video content, the videos collected for the \textit{VQA\textsuperscript{2} Instruction Dataset} are sourced from various image/video datasets \cite{lin2019kadid,hosu2020koniq,sinno2018large,ying2021patch,ghadiyaram2017capture,duanmu2016quality,duanmu2018quality,bampis2021towards,he2024videoscore}, providing a wide range of visual content and quality variations. We determine the sampling proportion of different quality levels according to each dataset's original quality distribution. This is because almost all source datasets exhibit a normal distribution of quality, and uniform sampling fails to capture sufficient videos across all quality levels. More importantly, we believe that moderate-quality videos encompass both positive and negative attributes, offering greater annotation value than videos of purely high or low quality, and this sampling strategy can guarantee sufficient medium-quality videos to be selected. Tab. \ref{tab:datainformation} summarizes our dataset's sampling and statistic information.

\begin{table}[!t]\huge
    \centering
\renewcommand\arraystretch{1.38}
\renewcommand\tabcolsep{2.5pt}
\belowrulesep=0pt\aboverulesep=0pt
    \caption{Source datasets, sampling information and statistic summary of the \textit{VQA\textsuperscript{2} Instruction Dataset}. }
    \vspace{-5pt}
   \resizebox{0.97\linewidth}{!}{\begin{tabular}{c:c:c:c:c}
    \hline
    \textbf{Stages} &\textbf{Data Types} &\textbf{Source Dataset} &\textbf{Original~/~Sampled \#} & \textbf{Instruction Pairs}  \\ \hline
    \multirow{4}{*}{\textit{Stage-1}}&\multirow{2}{*}{Images}&
     KonIQ-10k \cite{hosu2020koniq}& 10,073~/~3,693 &\multirow{2}{*}{7,174} \\
     &&KADID-10k \cite{lin2019kadid}& 10,125~/~3,481 \\ \cdashline{2-5}
     &\multirow{2}{*}{UGC-videos}&LIVE-Qualcomm \cite{ghadiyaram2017capture}& 208~/~34 &\multirow{2}{*}{5,211} \\ 
      &&LSVQ (train) \cite{ying2021patch} & 28,056~/~100 &\\ \hline
     \multirow{4}{*}{\textit{Stage-2}}&\multirow{1}{*}{UGC-videos}&
     LSVQ (train) & 28,056~/~28,056 &\multirow{1}{*}{28,056} \\\cdashline{2-5}
     &\multirow{3}{*}{Streaming-videos}&Waterloo-I \cite{duanmu2016quality}& 180~/~180 & \multirow{3}{*}{2,100}\\
     &&Waterloo-III \cite{duanmu2018quality}& 450~/~450 & \\ 
     &&LIVE-NFLX-II \cite{bampis2021towards}& 420~/~420 & \\
     \hline
     \multirow{4}{*}{\textit{Stage-3}}&\multirow{3}{*}{UGC-videos}&
     LSVQ (train)& 28,056~/~13,007 &\multirow{3}{*}{110,232} \\
     &&LSVQ (1080p) \cite{ying2021patch}& 3,573~/~998\\
     &&LIVE-VQC \cite{sinno2018large}& 585~/~497 & \\\cdashline{2-5}
     &\multirow{1}{*}{AIGC-videos}&VideoFeedback \cite{he2024videoscore}& 32,901~/~998 & 4,982 \\ \hline
     \multirow{2}{*}{\textit{Overall}} &\multicolumn{3}{c:}{Images: 7,174 / UGC videos (no overlap): 29,585}&\multirow{2}{*}{157,755} \\ &\multicolumn{3}{c:}{Streaming videos: 1,050 / AIGC videos: 998}&\\
    \hline
\end{tabular}
 }
\vspace{-8pt}
\label{tab:datainformation}
\end{table}
\subsection{Distortion-recognition Based Pretraining Set}
\label{Stage1}
Many classic multi-modal works \cite{radford2021learning,li2023blip,liu2024visual,ye2024mplug} follow the well-established pretraining-finetuning paradigm, which has been proven to be an effective way for foundation model development.
Since we believe that distortion is central to low-level quality assessment, and the recognition of distortion types is fundamental to achieving high performance on such tasks, we design \textit{VQA\textsuperscript{2} Instruction Dataset Stage-1}, a distortion recognition centered instruction subset, as the model's pre-training instruction set.

Our distortion recognition design includes both spatial and motion aspects. We sample $7,174$ images from the KonIQ-10K \cite{hosu2020koniq} and KADID-10K \cite{lin2019kadid} datasets for spatial distortion recognition. Using the distortion type annotations from \cite{zhang2023blind}, we select $11$ different types of spatial distortions: \textit{``compression artifact"}, \textit{``spatial blur"}, \textit{``motion blur"}, \textit{``noise"}, \textit{``overexposure"}, \textit{``underexposure"}, \textit{``low contrast"}, \textit{``high contrast"}, \textit{``oversaturation"}, \textit{``desaturation"} and \textit{``block effect"}.
For motion distortion recognition, we focus on the video distortions—\textit{``flickers (camera shake)" and ``stuttering"}. We use $34$ videos with flicker distortion from the LIVE-Qualcomm \cite{ghadiyaram2017capture} and $100$ videos with flicker and stuttering distortions in the LSVQ (train) \cite{ying2021patch} and extend them by extracting temporal and spatial clips, yielding $5,211$ video clips containing motion distortions. The format of instructions is shown in Fig. \ref{fig:stage12}.
\vspace{-10pt}
\subsection{Instruction Set for Quality Scoring}
\label{stage2}
Video quality scoring has always been a key focus of video quality assessment.
The accurate assessment of video quality forms the foundation for understanding and analyzing quality attributes within the video. To this end, we design the \textit{VQA\textsuperscript{2}  Instruction Dataset Stage-2}, a finetuning subset primarily aiming at scoring the video quality.
We use the LSVQ (train) as the video source for the off-line UGC video quality scoring task while using Waterloo-I \cite{duanmu2016quality}, Waterloo-III \cite{duanmu2016quality}, and LIVE-NFLX-II \cite{bampis2021towards} as the video sources for the streaming video quality scoring task. To ensure that the subjective experimental scores from all datasets are on the same scale, we normalize the MOSs in each dataset to the $[0,100]$ range. After scaling, we transform the video quality into five quality levels: \textit{``High"}, \textit{``Good"}, \textit{``Fair"}, \textit{``Poor"} and \textit{``Low"}, with each level representing a $20$-point interval. This approach minimizes the impact of inconsistent quality distributions across datasets.


Since rebuffering and long-term stalling are the major distortions in the streaming video datasets, we add \textbf{stalling information} in the instruction set. We design two formats to present the stalling information. The first format uses a \textit{``0/1"} sequence to directly indicate stalling for each frame (``\textit{1}" represents stalling, and ``\textit{0}" represents smooth playback). In the second format, the \textbf{stalling information} is summarized as follows: \textit{the total number of stalling events}, \textit{the duration of each stalling event}, \textit{the proportion of stalling events duration to the total video length}, \textit{the initial buffering time}, and \textit{the time elapsed between the end of the last stalling event and the end of the playback}. The UGC and streaming video prompt formats are shown in Fig. \ref{fig:stage12}.


\begin{figure*}[ht]
  \centering
  \includegraphics[width=0.97\linewidth]{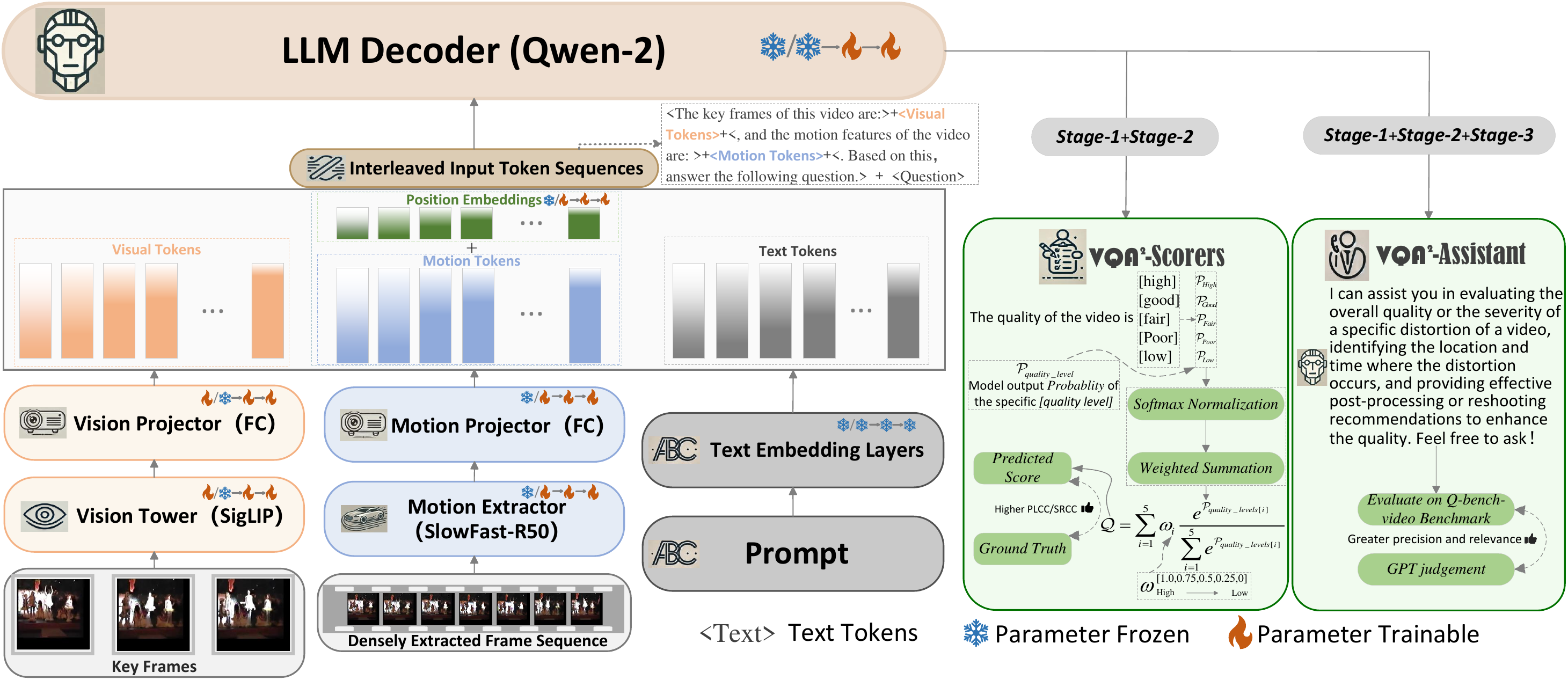}
    \vspace{-6pt}
   \caption{The model structure and training strategy. The model is based on the vanilla \textit{LLaVA-OneVision-Chat} and \textit{SlowFast-R50}. The training strategy (freeze or unfreeze) across the $3$ stages is separated by ``$\rightarrow$". Specifically, ``/" denotes the training strategy used during the ``image data training/video data training" in \textit{Stage-1}.
}
\vspace{-10pt}
   \label{fig: structure}
 
\end{figure*}

\subsection{Instruction Set for Quality Understanding}
\label{annotation}
The primary part of the dataset is  \textit{VQA\textsuperscript{2} Instruction Dataset Stage-3}, an instruction subset for low-level video visual quality understanding and question answering. We select $14,005$ videos from LSVQ (train) and LSVQ (1080p)
\cite{ying2021patch}, $497$ videos from the LIVQ-VQC \cite{sinno2018large}, and $998$ videos from the AIGC video dataset Videofeedback \cite{he2024videoscore}. The instruction set focuses on low-level visual quality question-answering for videos while also involving a small number of data related to video aesthetic assessment (VAA) and AIGC video quality analysis. For constructing more question-answer (Q\&A) pairs, we adopt the method of extending human expert annotations using \textit{GPT}.

\paragraph{Human Annotation Process.} The human annotation process for each video is divided into two parts. First, we require a comprehensive \textbf{overall quality depiction} focused on video quality attributes. Each depiction includes several quality attributes described with three key elements: 
 {\textbf{Quality Attribute}}$+${\textbf{Degree}}$+${\textbf{Temporal Description}}. A detailed example is shown in Fig. \ref{fig:stage3}. 
This design highlights the key point that the viewer perceives the video quality by focusing on 
salient quality attributes.  Furthermore, the depiction incorporates rich temporal information, significantly enhancing the model’s capacity to answer questions about the temporal quality within videos. We instruct annotators to select quality attributes for overall depiction according to the transformed video quality levels.
To maximize the value of medium-quality videos, annotators are encouraged to identify a broader range of quality attributes in these videos, encompassing both positive and negative aspects. 

In the second part, we require annotators to provide a brief quality depiction containing in-context (local) temporal or spatial quality. If such a depiction can not be found, it may be substituted with extended conversations, such as designing a Q\&A pair about possible causes of certain distortions or proposing feasible solutions to enhance video quality. The supplementary materials provide more detailed information for annotation experiments and examples.

\paragraph{GPT Extension.} Most of our Q\&A pairs are generated using \textit{GPT} to rewrite and refine the annotated overall quality depictions, in-context depictions, and extended conversations. For each overall depiction, we instruct \textit{GPT} to extract the information and reformulate it into three quality (attributes)-centric Q\&A pairs: a binary-choice single-answer pair, a multiple-choice single-answer pair, and an open-ended pair. Additionally, we require \textit{GPT} to generate one temporal-centric Q\&A pair and another extended conversation (except for the human-annotated one) based on the original overall depiction. For the human-annotated in-context depictions and extended conversations, the \textit{GPT} also rewrites them to the format of formal Q\&A pairs with their meaning unchanged.

\paragraph{Quality Causal Analysis.} For the annotated overall depictions mentioned before, to fully utilize the information during training, we manually formulate them into the form of quality causal analysis Q\&A pairs with the specific format shown in Fig. \ref{fig:stage3}.


\section{The VQA\textsuperscript{2} Series Models}
We propose the \textbf{\textit{VQA\textsuperscript{2} series models}} and a unique training pipeline. 
The structure of the model is illustrated in Fig. \ref{fig: structure}. The loss function in all three training stages adopts the standard generation loss used in text generation tasks, such as GPT-loss \cite{radford2019language}.

\subsection{Model Structure}
\paragraph{Base Model.}
We select \textit{LLaVA-OneVision-Chat-7B} \cite{li2024llava} as the foundation of our model. The foundation model achieves excellent performance on multiple high-level visual question answering benchmarks \cite{fu2024video,longpre2023flan} and demonstrates outstanding capabilities in video semantic understanding and reasoning. The foundation model includes: a vision tower constructed from the SigLIP \cite{zhai2023sigmoid}, which is used to extract feature tokens from keyframe sequences; a vision projector composed of fully connected layers for feature mapping; and the Qwen-2 model \cite{yang2024qwen2} with its tokenizer serving as the LLM and text embedding layers.

\paragraph{Motion Extractor and Motion Projector.}
We observe that many video LMMs  \cite{li2024llava,ye2024mplug1,chen2024far} can achieve excellent performance by only inputting keyframe sequences extracted at sparse sampling rates (\textit{1 frame per second, etc.}) when evaluated on high-level visual question answering tasks. We believe this is due to the high redundancy of temporal semantic information in videos. The semantic content in adjacent video frames is highly consistent, and changes or connections in video contents over longer periods can be effectively captured through sparsely sampled keyframe sequences.
However, this redundancy is sometimes almost nonexistent in video quality assessment. For example, video stuttering or shaking occurring within a short time can significantly degrade a video's perception quality, and such distortion is closely related to the adjacent frames. Therefore, keyframe sequences extracted at long intervals will completely miss this distortion representation.

In summary, we propose that the model should incorporate a video motion extraction module that processes adjacent frames from densely input video frames. 
We select the SlowFast-R50 \cite{feichtenhofer2019slowfast} for motion extraction, inputting the entire video after spatial pre-processing. 
To ensure that the number of the motion tokens aligns with the number of frames of the video, we set $\tau=4$\footnote{$\tau$ denotes the sampling interval (frames) in the slow path.} and $\alpha=4$ \footnote{$\alpha$ denotes the ratio of the sampling intervals of the slow and fast path.} and use only the fast path features as the motion tokens. We then employ a motion projector identical in structure to the vision projector to map the motion tokens, ensuring their dimensions are consistent with the visual tokens and text tokens. Additionally,  we apply positional encoding by performing token-wise addition on the motion tokens and learnable absolute positional embeddings. During training, the visual tokens sequence and motion tokens sequence are input in an interleaved manner. 

\subsection{Distortion-recognition Based Pretraining}
We first pre-train the model using the data from \textit{Stage-1}. When training with the spatial distortion instruction subset, we freeze the LLM, motion extractor, and motion projector, allowing only the vision tower and vision projector to be trained. In contrast, when using the motion distortion instruction subset, we exclusively train the motion extractor and projector. This is because we think the pre-training data has a relatively simple format. Unfreezing all model parameters for training could easily lead to severe overfitting, thereby affecting the subsequent training process. 

\subsection{The VQA\textsuperscript{2}-Scorers and the VQA\textsuperscript{2}-Assistant}
\label{S/Amodels}
The \textit{VQA\textsuperscript{2}-Scorers} consist of the \textit{VQA\textsuperscript{2}-UGC-Scorer} and the \textit{VQA\textsuperscript{2}-Streaming-Scorer}, which are specially developed for scoring the quality of UGC and streaming videos, respectively. The \textit{VQA\textsuperscript{2}-UGC-Scorer} is trained on the instruction subset from the \textit{Stage-2} UGC video data portion upon the pre-trained model. Subsequently, the \textit{VQA\textsuperscript{2}-Streaming-Scorer} is further trained on the instruction set from the \textit{Stage-2} streaming video data portion based on the \textit{VQA\textsuperscript{2}-UGC-Scorer}. The reason for this setting is given in the supplementary materials.
The scoring methodology for model inference, referenced from \cite{wu2023q}, is shown in Fig. \ref{fig: structure} and detailed in the supplementary materials. At this stage, all training involves full parameter tuning. 

The final model, the \textit{VQA\textsuperscript{2}-Assistant}, is designed to master more nuanced video quality understanding tasks and efficient low-level visual question answering while still possessing the capabilities for precise quality scoring. Building on the \textit{VQA\textsuperscript{2}-UGC-Scorer}, this model also undergoes full parameter tuning using the instruction subset \textit{Stage-3}.

\section{Experiments}


We conduct comprehensive experiments on video quality scoring and understanding tasks to validate the performance of our model family. Additionally, we perform detailed ablation studies to analyze the impact of various attributes.

\subsection{Experimental Setups}
\label{setup}
\paragraph{Training Strategies.}  We strictly follow the training  hyperparameters provided in the \textit{LLaVA-OneVision} project. All the models are trained for only one complete epoch on their respective training data. 

\paragraph{Prompt Design.} 
During the training and evaluation stages, a system prompt is added before all instructions in the dataset. We design the system prompt to be identical in all stages of training. 
During the evaluation, we employ specially designed system prompts based on the specific task. To thoroughly verify the model's end-to-end performance for the video quality scoring tasks, we do not provide any time information (like length, frame rate, and stalling information) that may need additional extraction about the test video in the system prompt.
In the video quality understanding task, to meet the format requirements of the evaluation benchmark,
we add time information to the evaluation prompt, including the number of video frames, sampled keyframes, and the sampling interval. 
The specific system prompt formats in the training and evaluation stages are presented in the supplementary materials.

\subsection{Evaluation on Quality Scoring Tasks}
We assess the quantitative scoring capabilities of \textit{VQA\textsuperscript{2}-UGC-Scorer} and \textit{VQA\textsuperscript{2}-Assistant} on $4$ open-source UGC-VQA datasets \cite{ying2021patch,hosu2017konstanz,sinno2018large,wang2019youtube}, while evaluating the  \textit{VQA\textsuperscript{2}-Streaming-Scorer} on the Waterloo-IV \cite{duanmu2020waterloo}, the largest streaming-VQA dataset with $1,350$ videos. 
We select some vanilla LMMs \cite{ye2024mplug,liu2024visual} and our base model (with vanilla \textit{LLaVA-ov-chat 
(7B)} and \textit{SlowFast-R50}) for reference. Additionally, we choose several high-performing UGC-VQA models (\textit{FAST-VQA} \cite{wu2022fast}, \textit{Minimalistic-VQA} \cite{sun2024analysis}, \textit{DOVER} \cite{wu2023exploring}, \textit{Modular-VQA} \cite{wen2024modular}, \textit{q-align} \cite{wu2023q}, and \textit{q-instruct} \cite{wu2024q}) and Streaming-VQA models (\textit{SQI} \cite{duanmu2016quality}, \textit{BSQI} \cite{duanmu2023bayesian}, and \textit{DSA-QoE} \cite{jia2024continuous}) for further comparison. 

We use \textit{Pearson Linear Correlation Coefficient} (PLCC) and  \textit{Spearman Rank Correlation Coefficient} (SRCC) as evaluation metrics. Apart from the \textit{q-align-IQA}, \textit{q-align-onealign}, \textit{q-instruct}, \textit{VQA\textsuperscript{2}-Streaming-Scorer}, and \textit{VQA\textsuperscript{2}-Assistant} which have specific training sets and the training-free method \textit{SQI}, all models in both tasks are trained or optimized on the same datasets (LSVQ (train) for UGC video quality scoring / Waterloo-I, Waterloo-III, and LIVE-NFLX-II for streaming video quality scoring). Specifically, since \textit{Stage-3} involves videos from the LIVE-VQC and LSVQ (1080p), we remove these videos to avoid leakage in all experiments, leaving $88$ videos out of $585$ in LIVE-VQC and $2,575$ videos out of $3,573$ in LSVQ (1080p) for evaluation. Tabs. \ref{tab:assessment} and \ref{tab:streaming} present the performance of our models and the comparison models on UGC / streaming video quality scoring tasks, respectively.  
\begin{table}[t]\huge
    \centering
    \renewcommand\arraystretch{1.25}
    \renewcommand\tabcolsep{4pt}
    \belowrulesep=0pt\aboverulesep=0pt

    \caption{Performance (SRCC$\uparrow$~/~PLCC$\uparrow$) on UGC video quality scoring tasks. The best result is denoted in \textbf{bold},
the second best is \underline{underlined}, and the third best is
in \textit{italic} font. }
        \vspace{-5pt}
    \resizebox{1\linewidth}{!}
    {\begin{tabular}{l|cc|ccc}
 \hline
    \multicolumn{1}{l|}{\textbf{Dataset Scale}} &  \multicolumn{2}{c|}{\textbf{Large-scale}} & \multicolumn{3}{c}{{\textbf{Medium~/~Small-scale}}}\\ \cdashline{1-6}
      \multicolumn{1}{l|}{\textit{Test video \#}} &  2,575 & 7,182&88&1,380 &1,200 \\  \cdashline{1-6}
     \multicolumn{1}{l|}{\textit{Models}}&{\textit{LSVQ(1080p)}} & {\textit{LSVQ(test)}} & {\textit{LIVE-VQC}} &  {\textit{YouTube-UGC}}&\textit{KoNViD-1k} \\ \hline 
      \multicolumn{1}{l}{\textit{Classic}}\\ 
      \cdashline{1-6}
      \textit{FAST-VQA}  &  0.765~/~0.810 & 0.874~/~0.878 & 0.769~/~0.815& 0.725~/~0.742 & 0.859~/~0.857\\
     \textit{Minimalist-VQA} & 0.769~/~0.818 &0.880~/~0.872 & 0.765~/~0.812 & 0.783~/~0.799 & 0.859~/~0.861 \\
     \textit{Dover} & \underline{0.787}~/~0.828 &\textit{0.888}~/~\underline{0.886} & 0.771~/~0.819& 0.801~/~0.814 & \underline{0.890}~/~\textit{0.883}\\
     \textit{Modular-VQA} & \textbf{0.791}~/~\underline{0.844} &\underline{0.894}~/~\textbf{0.891} & 0.783~/~\textit{0.825} & 0.786~/~0.803& 0.878~/~\textbf{0.884}\\
     \cdashline{1-6}
      \multicolumn{1}{l}{\textit{LMMs (\textit{7B})}}\\ 
      \cdashline{1-6}
      \textit{mPLUG-Owl-2}  &  0.398~/~0.422 & 0.422~/~0.434 & 0.450~/~0.459& 0.437~/~0.448 & 0.532~/~0.532\\
     \textit{LLaVA-v1.5} & 0.341~/~0.355 &0.441~/~0.412 &0.242~/~0.302& 0.356~/~0.378 & {0.435}~/~0.419 \\
     \cdashline{1-6}
    \multicolumn{1}{l}{\textit{q-align (\textit{7B})} }\\ 
    \cdashline{1-6}
    \textit{IQA} &0.764~/~\textit{0.842}&0.729~/~0.733 & 0.724~/~0.772 & 0.715~/~0.723 & 0.797~/~0.780\\
     \textit{VQA}& 0.758~/~0.833 & 0.883~/~0.882 & \underline{0.777}~/~0.813 &0.811~/~0.830& 0.865~/~0.876\\
     \textit{onealign}& 0.761~/~0.822 &0.886~/~0.884 &0.766~/~\underline{0.826}&\underline{0.831}~/~\textbf{0.847} &0.876~/~0.878 \\
     \cdashline{1-6}
     \multicolumn{1}{l}{\textit{q-instruct (\textit{7B})} }\\
     \cdashline{1-6}\textit{mPLUG-Owl-2} & 0.602~/~0.580 & 0.644~/~0.640 & 0.660~/~0.673& 0.601~/~0.622 &0.492~/~0.520 \\
     \textit{LLaVA-v1.5}& 0.571~/~0.562 & 0.610~/~0.578 & 0.685~/~0.616& 0.635~/~0.667 & {0.664}~/~0.577\\
     \cdashline{1-6}
     \multicolumn{1}{l}{\textit{VQA\textsuperscript{2} (\textit{7B})}}\\
     \cdashline{1-6}
     \textit{base} & 0.527~/~0.609 & 0.658~/~0.638 & 0.665~/~0.751& 0.626~/~0.631 & {0.692}~/~0.691  \\
     \textit{UGC-scorer}& \textit{0.782}~/~\textbf{0.847} & \textbf{0.897}~/~\textit{0.885} & \textbf{0.785}~/~\textbf{0.830} & \textit{0.814}~/~\textit{0.832}&\textbf{0.894}~/~\textbf{0.884} \\
      \textit{Assistant}& 0.760~/~0.819 & 0.882~/~0.856 & \textit{0.776}~/~0.823 & \textbf{0.854}~/~\underline{0.841}& \textit{0.883}~/~0.844 \\
  \hline
    \end{tabular}}
      \vspace{-7pt}
    \label{tab:assessment}
\end{table}

\begin{table}[t]\huge
    \centering
    \renewcommand\arraystretch{1.23}
    \renewcommand\tabcolsep{2.5pt}
    \belowrulesep=0pt\aboverulesep=0pt
    
    \caption{Performance (SRCC$\uparrow$~/~PLCC$\uparrow$) on streaming video quality scoring tasks. \textit{Type 1-5} refers to the $5$ unique video content types included in the Waterloo-IV dataset: \textit{game, documentary, movie, nature, and sports}, each comprising $270$ videos. \textit{All} refers to the entire dataset. For better visualization, all results are rounded to two decimal places. }
    \vspace{-5pt}
    \resizebox{\linewidth}{!}
    {\begin{tabular}{l|ccccc|c}
     \hline
     \multicolumn{1}{l|}{\textbf{Dataset}}&  \multicolumn{6}{c}{\textbf{Waterloo-IV}}\\  \cdashline{1-7}
     \textit{Models} &\textit{Type 1} &\textit{Type 2}&\textit{Type 3}&\textit{Type 4}&\textit{Type 5}&\textit{All}\\
     \cdashline{1-7}
      \multicolumn{1}{l}{\textit{Classsic}}\\
      \cdashline{1-7}
      \textit{SQI}  &0.77~/~0.79&0.75~/~0.76&0.68~/~0.70&0.64~/~0.62&0.76~/~0.78&0.67~/~0.67\\
     \textit{BSQI} &0.79~/~0.81&\textit{0.75}~/~\underline{0.79}&0.70~/~\textit{0.72}&\textit{0.68}~/~\textbf{0.72}&0.80~/~0.78&\textit{0.70}~/~\textit{0.72}\\
     \textit{DSA-QoE} &\underline{0.80}~/~0.81&\textbf{0.76}~/~\textbf{0.79}&\underline{0.71}~/~\underline{0.72}&0.68~/~\underline{0.71}&\textit{0.81}~/~0.80&\underline{0.73}~/~\underline{0.74}\\
     \cdashline{1-7}
      \multicolumn{1}{l}{\textit{LMMs (\textit{7B})}}\\
      \cdashline{1-7}
     \textit{LLaVA-v1.5}&0.54~/~0.45&0.55~/~0.61&0.45~/~0.46&0.38~/~0.42&0.75~/~0.81&0.33~/~0.38\\
     \cdashline{1-7}
    \multicolumn{1}{l}{\textit{q-align (\textit{7B})}}\\
    \cdashline{1-7}
    \textit{IQA} &0.68~/~0.77& 0.66~/~0.69 & 0.49~/~0.43 & \underline{0.69}~/~\textit{0.71}&0.80~/~\textit{0.81}&0.36~/~0.31\\
     \textit{VQA}&\underline{0.80}~/~\textit{0.85}&0.68~/~0.76&0.56~/~0.70&0.66~/~0.63&0.78~/~\underline{0.82}&0.46~/~0.50\\
     
     \textit{onealign}&0.77~/~\underline{0.86}&0.73~/~0.78&\textit{0.70}~/~0.59&0.49~/~0.56&\textbf{0.86}~/~0.80&0.43~/~0.38\\
     \cdashline{1-7}
     \multicolumn{1}{l}{\textit{VQA\textsuperscript{2} (\textit{7B})}}\\
     \cdashline{1-7}
     \textit{base}&0.32~/~0.37& 0.67~/~0.69 & 0.49~/~0.43 & 0.39~/~0.42&0.74~/~0.76&0.37~/~0.37\\
    \textit{Streaming-scorer }&\textbf{0.85}~/~\textbf{0.90}& \underline{0.75}~/~\textit{0.78} & \textbf{0.72}~/~\textbf{0.74} & \textbf{0.70}~/~0.68&\underline{0.86}~/~\textbf{0.82}&\textbf{0.76}~/~\textbf{0.78}\\
    \hline
    \end{tabular}}
    \vspace{-10pt}
    \label{tab:streaming}
\end{table}

\begin{table}[t]\large
    \centering
    \renewcommand\arraystretch{1.46}
    \renewcommand\tabcolsep{2.8pt}
    \belowrulesep=0pt\aboverulesep=0pt

    \caption{Evaluation results on the {\textit{test}} subset of the Q-bench-video. We use \textcolor{red}{red} font to indicate the performance improvement of the
trained model compared to the base model.}
 \vspace{-5pt}
    \resizebox{\linewidth}{!}{\begin{tabular}{l|ccc|ccc|c}
     \hline
        \textbf{Categories} & \multicolumn{3}{|c|}{\textbf{Question Types}} & \multicolumn{3}{c|}{\textbf{Quality Concerns}} & \multirow{2}{*}{{\textit{Overall$\uparrow$}}} \\ \cdashline{1-7}
        \multirow{1}{*}{\textit{LMMs} }  & {\textit{Binary$\uparrow$}}& {\textit{Multi.$\uparrow$}} & {\textit{Open$\uparrow$}} & \multirow{1}{*}{\textit{Tech.$\uparrow$}}& \multirow{1}{*}{\textit{Temp.$\uparrow$}}  &\multirow{1}{*}{\textit{Other}$\uparrow$} \\
      \hline
        \multicolumn{8}{l}{\textit{Image LMMs} (\textit{7B})} \\ \hdashline
         \textit{mPLUG-Owl-2} & 57.72\% & 42.61\% & 32.39\% & 41.99\% & 46.46\% & 44.79\% & 43.99\% \\
        \textit{LLaVA-v1.5} & 58.39\% & 50.17\% & 39.78\% & 46.31\% & 48.32\% & 54.01\% & 49.23\% \\
        \hdashline
        \multicolumn{8}{l}{\textit{Video LMMs} (\textit{7B})} \\ \hdashline
        \textit{mPLUG-Owl3} & 58.05\% & \textit{57.73\%} & 38.99\% & \textit{53.04\%} & \underline{51.68\%} & 51.34\% & 51.27\%\\
        \textit{LLaVA-ov-chat}   & 57.05\% & 53.95\% & 33.65\% & 47.60\% & 45.45\% & 51.07\% & {47.85\%} \\
        \textit{InternVL-Chat}& \underline{65.77\%} & 55.33\% & 34.43\% & 47.04\% & 50.34\% & \textbf{57.75\%} & \textit{51.43\%} \\
        \textit{VILA1.5} (\textit{7B})  & 58.05\% & 45.36\% & 37.26\% & 45.99\% & 45.96\% & {49.33\%} & 46.69\% \\
        \textit{LLaVA-Next-Video}  & \textit{65.10\%} & 50.17\% & 38.21\% & 48.80\% & 48.15\% & \textit{{55.75\%}} & 50.88\% \\
        \hdashline
        \multicolumn{8}{l}{\textit{GPT-Series} } \\ \hdashline
         \textit{GPT-4V} &58.25\% & 50.17\% & \textbf{44.62\%} & 51.38\% & 47.79\% & 51.61\% & 50.89\% \\
        \textit{GPT-4o-mini} & 50.17\% & 45.99\% & 37.34\% & 44.98\% & 46.60\% & 33.85\% & 44.33\% \\
         \textit{GPT-4o}  & 58.91\% & \underline{57.80\%} & \underline{43.97\%} & \underline{53.24\%} & \textit{50.79\%} & \underline{56.92\%} & \underline{53.32\%}\\
         \hdashline
        \multicolumn{8}{l}{\textit{VQA\textsuperscript{2} (\textit{7B})}} \\ \hdashline
         \textit{base} &52.86\% &53.31\% &34.65\% &46.68\% &46.77\% &50.13\% &46.61\%\\
         \multirow{2}{*}{\textit{Assistant }}  & \textbf{67.79\%} & \textbf{61.86\%} & \textit{41.82\%} & \textbf{56.73\%}& \textbf{57.41\%}& 51.34\%& \textbf{56.78\%} \\
         &$\text{\textcolor{red}{+14.93\%}}$&$\text{\textcolor{red}{+8.55\%}}$&$\text{\textcolor{red}{+7.17\%}}$&$\text{\textcolor{red}{+10.05\%}}$&$\text{\textcolor{red}{+10.64\%}}$ &$\text{\textcolor{red}{+1.21\%}}$ &$\text{\textcolor{red}{+10.17\%}}$\\
        \hline
    \end{tabular}}
    \vspace{-8pt}
    \label{tab:sub_test}
\end{table}

\begin{table}[t]\large
    \centering
    \renewcommand\arraystretch{1.46}
    \renewcommand\tabcolsep{2.8pt}
    \belowrulesep=0pt\aboverulesep=0pt
     
    \caption{Evaluation results on the {\textit{dev}} subset of the Q-bench-video. }
  \vspace{-8pt}
    \resizebox{\linewidth}{!}{
    \begin{tabular}{l|ccc|ccc|c}
     \hline
        \textbf{Categories} & \multicolumn{3}{|c|}{\textbf{Question Types}} & \multicolumn{3}{c|}{\textbf{Quality Concerns}} & \multirow{2}{*}{{\textit{Overall$\uparrow$}}} \\ \cdashline{1-7}
        \multirow{1}{*}{\textit{LMMs}}& {\textit{Binary$\uparrow$}}& {\textit{Multi.$\uparrow$}} & {\textit{Open$\uparrow$}} & \multirow{1}{*}{\textit{Tech.$\uparrow$}}& \multirow{1}{*}{\textit{Temp.$\uparrow$}}  &\multirow{1}{*}{\textit{Other}$\uparrow$} \\
      \hline
        \multicolumn{8}{l}{\textit{Image LMMs} (\textit{7B})} \\ \hdashline
        \textit{mPLUG-Owl-2} & 61.79\% & 37.35\% & 33.39\% & 42.67\% & 51.94\% & 45.66\% & 44.27\% \\
        \textit{LLaVA-v1.5}&63.79\% & 46.99\% & 36.31\% & 49.33\% & 45.58\% & 48.51\% & 49.34\%\\
        \hdashline
        \multicolumn{8}{l}{\textit{Video LMMs} (\textit{7B})} \\ \hdashline
        \textit{mPLUG-Owl3}  & 59.14\% & \textbf{57.23\%} & 38.50\% & 52.00\% & 54.77\% & \textbf{51.76\%} & 52.21\%\\
        \textit{LLaVA-ov-chat }  & 60.47\% & \textit{53.01\%} & 32.85\% & 49.42\% & 52.12\% & 45.81\% & 49.39\% \\
        \textit{InternVL-Chat} & \underline{70.43\%} & 49.70\% & 33.39\% & 50.25\% & 51.41\% & 49.59\% & 51.65\% \\
        \textit{VILA1.5} & 58.05\% & 45.36\% & 37.26\% & 45.99\% & 45.96\% & 49.33\% & 46.69\% \\
        \textit{LLaVA-Next-Video} & \textit{69.77\%} & 44.58\% & 33.39\% & 49.08\% & 49.65\% & \textit{50.00\%} & 49.56\% \\
        \hline
        \multicolumn{8}{l}{\textit{GPT-Series}}  \\ \hdashline
         \textit{GPT-4V} &64.78\% & 51.20\% & \textbf{43.43\%} & \underline{55.25\%} & \textit{56.18\%} & 50.00\% & \textit{53.36\%}  \\
         \textit{GPT-4o-mini} &54.49\% & 43.98\% & \textit{38.87\%} & 45.58\% & 54.42\% & 47.15\% & 45.92\% \\
         \textit{GPT-4o} & 66.01\% & 52.56\% & \underline{42.78\%} & \textit{52.21\%} & \textbf{62.59\%} & \underline{50.28\%} & \underline{54.07\%} \\
         \hline
        \multicolumn{8}{l}{\textit{VQA\textsuperscript{2} (\textit{7B})}} \\ \hdashline
         \textit{base } &55.78\% & 45.43\% & 31.48\% & 45.25\% & 50.73\% & 44.69\% & 44.62\% \\
         \multirow{2}{*}{\textit{Assistant}}  & \textbf{75.42\%} & \underline{56.93\%} & 37.78\% & \textbf{61.04\%} & \underline{60.36\%} & 46.87\% & \textbf{56.50\%}\\
         &$\text{\textcolor{red}{+19.64\%}}$&$\text{\textcolor{red}{+11.50\%}}$&$\text{\textcolor{red}{+6.30\%}}$&$\text{\textcolor{red}{+15.79\%}}$&$\text{\textcolor{red}{+9.63\%}}$&$\text{\textcolor{red}{+2.18\%}}$&$\text{\textcolor{red}{+11.88\%}}$\\
        \hline
    \end{tabular}}
    \vspace{-10pt}
    \label{tab:sub_dev}
\end{table}
Experimental results demonstrate that the \textit{UGC-Scorer} and \textit{Streaming-Scorer} achieve the best performance across most datasets and metrics and rank within the top $3$ in almost all of them. 
Although the \textit{Assistant} slightly lags behind the \textit{UGC-Scorer} in performance, it still delivers a relatively strong scoring performance. This confirms that the \textit{Assistant}, primarily designed for video quality understanding and question answering, can still effectively handle quality scoring tasks, showcasing its versatility.

\subsection{Evaluation on Quality Understanding Tasks}

We evaluate the capabilities of the \textit{VQA\textsuperscript{2}-Assistant} on video quality understanding tasks using the Q-Bench-Video \cite{zhang2024q}, a comprehensive benchmark for video quality understanding tasks for LMMs. It contains $1,800$ videos and $2,378$ multi-type questions. This evaluation encompasses the model's answer accuracy and relevance across $3$ question types: binary yes/no questions (\textit{Binary}), multiple-choice (single answer) questions (\textit{Multi.}), and open-ended questions (\textit{Open}). These questions also span $3$ quality concerns: video technical quality aspect (\textit{Tech.}), video temporal quality aspect (\textit{Temp.}), and other categories (\textit{Other}), which include AIGC and VAA, etc. The model's responses are evaluated using \textit{GPT} with the same settings in Q-Bench-Video. Specifically, since our training does not include multi-video analysis, we exclude the questions involving multi-video ($564$ out of $2,378$). We select a series of high-performing open-source video LMMs \cite{ye2024mplug,liu2024visual, ye2024mplug1,li2024llava,chen2024far,ke2023vila,zhang2024video} and proprietary \textit{GPT series} \cite{achiam2023gpt} for comparison. The keyframe sequence input for all models is sampled at $1$ frame per second.  Except for the \textit{GPT series}, we set the \textit{greedy search} scheme for model generation, ensuring all results are reproducible.  The accuracy  of all models on the \textit{test} and \textit{dev} sets of the Q-Bench-Video are presented in Tabs. 
\ref{tab:sub_test} and \ref{tab:sub_dev}.

Experimental results indicate that the \textit{Assistant} outperforms the base model across all sub-dimensions in both subsets. In terms of question types, the \textit{Assistant} achieves the most significant improvement on binary questions, surpassing the base model by \textit{\textcolor{red}{$19.64\%$}} and \textit{GPT-4o} by \textit{\textcolor{red}{$8.88\%$}} on the \textit{test} set; and it also shows notable gains on what/how questions. As for quality concerns, the \textit{Assistant} achieves substantial improvements in the \textit{Tech.} and \textit{Temp.} sub-dimensions, exceeding the base model by \textit{\textcolor{red}{$10.05\%$}} and \textit{\textcolor{red}{$10.64\%$}} in each on the \textit{test} set.  This underscores the importance of centering human annotations on quality attributes while incorporating extensive temporal and motion descriptions. 
Finally, the \textit{Assistant} outperforms \textit{GPT-4o} in \textit{overall} scores on both the \textit{test} and \textit{dev} subsets.

\begin{table}[t]\small
    \centering
    \renewcommand\arraystretch{1.1}
    \renewcommand\tabcolsep{4pt}
    \belowrulesep=0pt\aboverulesep=0pt
    \caption{Comparison of the \textit{UGC-Scorer} performance with / without the \textit{Stage-1}. The better is in \textbf{bold}.}
    \vspace{-6pt}
    \resizebox{1\linewidth}{!}
    {\begin{tabular}{c|cc|cc}
    \hline
    \multicolumn{1}{c|}{\textbf{Experiment Types}} &  \multicolumn{2}{c|}{\textbf{Intra-dataset}} & \multicolumn{2}{c}{{\textbf{Cross-dataset}}}\\ \cdashline{1-5}
     \multicolumn{1}{c|}{\textbf{Models}}&{\textit{LSVQ(test)}} & {\textit{LSVQ(1080p)}} & {\textit{LIVE-VQC}} & \textit{KoNViD-1k} \\ \hline 
     \textit{wo-Stage1}& 0.887~/~ 0.879 &0.751~/~0.812  &0.776~/~0.819 & 0.884~/~0.880 \\
      \textit{w-Stage1}& \textbf{0.897}~/~\textbf{0.885} & \textbf{0.782}~/~\textbf{0.847} & \textbf{0.785}~/~\textbf{0.830} & \textbf{0.894}~/~\textbf{0.884} \\
    \hline
    \end{tabular}}
  \vspace{-6pt}
    \label{tab:Pre-training}
\end{table}
\vspace{-7pt}
\begin{table}[t]\large
    \centering
    \renewcommand\arraystretch{1.1}
    \renewcommand\tabcolsep{3pt}
    \belowrulesep=0pt\aboverulesep=0pt
    \caption{Comparison of models performance with and without the motion features extractor.}
    \vspace{-6pt}
    \resizebox{1\linewidth}{!}
    {\begin{tabular}{c|cc|ccc}
    \hline
    \multicolumn{1}{c|}{\textbf{Models}} &  \multicolumn{2}{c|}{\textbf{UGC-Scorer}} & \multicolumn{3}{c}{{\textbf{Assistant (\textit{dev} set)}}}\\ \cdashline{1-6}
     \multicolumn{1}{c|}{\textbf{Version}}& {\textit{LIVE-VQC}} & \textit{KoNViD-1k}&\textit{Tech.$\uparrow$}&\textit{Temp.$\uparrow$}&\textit{Overall$\uparrow$} \\ \hline 
     \textit{wo-Motion}& 0.773~/~0.822 & 0.873~/~0.865 & 58.17\% & 55.85\% & 54.30\% \\
     \textit{w-Motion}&\textbf{0.785}~/~\textbf{0.830} & \textbf{0.894}~/~\textbf{0.884} & \textbf{61.04\%}$_\text{\textcolor{red}{+2.87\%}}$ & \textbf{60.36\%}$_\text{\textcolor{red}{+4.51\%}}$ &\textbf{56.50\%}$_\text{\textcolor{red}{+2.20\%}}$ \\
    \hline
    \end{tabular}}
  \vspace{-6pt}
    \label{tab:Motion}
\end{table}
\begin{table}[t]\tiny
    \centering
    \renewcommand\arraystretch{1.1}
    \renewcommand\tabcolsep{5pt}
    \belowrulesep=0pt\aboverulesep=0pt
    \caption{Comparison of the \textit{UGC-Scorer} performance with different data mixture strategies.}
    \vspace{-6pt}
    \resizebox{1\linewidth}{!}
    {\begin{tabular}{c|cccc}
    \hline
     \multicolumn{1}{c|}{\textbf{Version}}&{\textit{LIVE-VQC}} & {\textit{LSVQ(1080p)}} & {\textit{LSVQ(test)}} & \textit{KoNViD-1k} \\ \hline 
     \textit{Mix}&\textbf{0.789}~/\textbf{0.836} & \textbf{0.761}~/~\textbf{0.831} & \textbf{0.883}~/~\textbf{0.873} & \textbf{0.895}/\textbf{0.892} \\
      \textit{Sequential}&0.776~/~0.823 & 0.760~/~0.819 & 0.882~/~0.856 & 0.883~/~0.844 \\
    \hline
    \end{tabular}}
     \vspace{-6pt}
    \label{tab:mix1}
\end{table}
\begin{table}[t]\small
    \centering
    
    \renewcommand\arraystretch{1.1}
    \renewcommand\tabcolsep{4pt}
    \belowrulesep=0pt\aboverulesep=0pt
    \caption{Comparison of the \textit{Assistant} performance on the \textit{test} set of Q-bench-video with different data mixture strategies.}
    \vspace{-6pt}
    \resizebox{\linewidth}{!}{\begin{tabular}{c|ccc|ccc|c}
    \hline
        
        \multirow{2}{*}{\textbf{Version}} & \multicolumn{3}{|c|}{\textbf{Question Types}} & \multicolumn{3}{c|}{\textbf{Quality Concerns}} & \multirow{2}{*}{{\textit{Overall$\uparrow$}}} \\ \cdashline{2-7}
        & {\textit{Binary$\uparrow$}}& {\textit{Multi.$\uparrow$}} & {\textit{Open$\uparrow$}} & \multirow{1}{*}{\textit{Tech.$\uparrow$}}& \multirow{1}{*}{\textit{Temp.$\uparrow$}}  &\multirow{1}{*}{\textit{Other}$\uparrow$} \\
      \hline
        \textit{Mix} &65.44\% & 57.04\% & 39.78\% & 56.25\% & 54.55\% & 47.19\% & 53.75\% \\
         \textit{Sequential}  & \textbf{67.79\%}& \textbf{61.86\%} & \textbf{41.82}\%& \textbf{56.73\%}& \textbf{57.41\%}& \textbf{51.34}\%&\textbf{56.78\%}\\
        \hline
         \textit{improvement} &$\text{\textcolor{red}{+2.35\%}}$ & $\text{\textcolor{red}{+4.82\%}}$& $\text{\textcolor{red}{+2.04\%}}$ & $\text{\textcolor{red}{+0.48\%}}$& $\text{\textcolor{red}{+2.86\%}}$ & $\text{\textcolor{red}{+4.15\%}}$ & $\text{\textcolor{red}{+3.03\%}}$  \\
       \hline
       \end{tabular}}
    \vspace{-10pt}
    \label{tab:mix2}
\end{table}

\vspace{6pt}
\subsection{Ablation Study}

We conduct several ablation studies and provide corresponding analyses.

\paragraph{\#1: Effects of Pre-training.}
We remove \textit{Stage-1} and directly train the \textit{UGC-Scorer} using the data from \textit{Stage-2}. Compared to the fully trained model, the resulting models' performance is shown in Tab. \ref{tab:Pre-training}. Experimental results indicate that the pretraining \textit{Stage-1} plays a crucial role throughout the training process.

\paragraph{\#2: Effects of Motion Extraction.}
We remove the motion extractor and projector, then follow the same training steps to obtain the \textit{Scorers} and \textit{Assistant} models. The models' performance on KonVID-1k and LIVE-VQC, as well as their scores on the \textit{Tech.}, \textit{Temp.}, and \textit{Overall} sub-dimensions in the Q-Bench-Video \textit{dev} subset, are presented in Tab. \ref{tab:Motion}. Experimental results show that motion feature extraction plays a significant role, especially when evaluating the \textit{Temp.} quality concern.
\begin{figure}[t]
  \centering
  \includegraphics[width=0.96\linewidth]{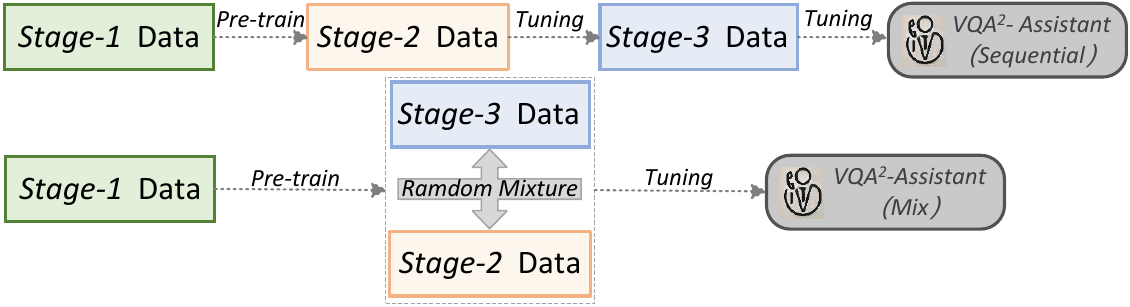}
    \vspace{-8pt}
   \caption{Two different data-combining strategies.}
   \label{fig: ablation}
\vspace{-9pt}
\end{figure}

\paragraph{\#3: Ablation Study on Different Data-combining Strategies.} During the training of the \textit{VQA\textsuperscript{2}-Assistant}, we alternatively combine the \textit{Stage-2} and \textit{Stage-3} data by randomly mixing them. The data-combining strategies are illustrated in Fig. \ref{fig: ablation}. We compare the performance of the \textit{Mix} and \textit{Sequential} versions of the model on the UGC video quality scoring task and the \textit{test} set of Q-Bench-Video, with the results reported in Tabs. \ref{tab:mix1} and \ref{tab:mix2}.

The model trained with the \textit{Mix} strategy further improves scoring performance in several datasets, suggesting that, beyond directly answering video quality-level questions, incorporating detailed quality understanding instructions still contributes to quantitative quality scoring. However, the \textit{Mix} version experiences a notable decline in quality understanding tasks. We attribute this to the scoring-related instructions in \textit{Stage-2}, which typically have a relatively simple focus and format. This impedes the
model from handling diverse question types effectively.

\section{Conclusion}
We introduce the \textit{VQA\textsuperscript{2} Instruction Dataset}, the first large-scale instruction dataset ever dedicated to video quality assessment through visual question answering, alongside the \textit{VQA\textsuperscript{2} series models} built on this dataset. The dataset construction spans $3$ stages and includes $157,755$ instruction pairs from diverse video types. Through comprehensive experiments on video quality scoring and quality understanding tasks, our models achieve excellent results in video quality scoring and outperform \textit{GPT-4o} in video quality understanding. The \textit{VQA\textsuperscript{2}-Assistant} excels in both quality scoring and understanding tasks, effectively fulfilling the demand for model versatility.

{
    \small
    \bibliographystyle{ieeenat_fullname}
    \bibliography{main}
}
\clearpage
\maketitlesupplementary
\appendix

\begin{table*}[h]
\belowrulesep=0pt\aboverulesep=0pt
\centering
\caption{
Details of the model structure and hyper-parameters for the model training.
}
\resizebox{\linewidth}{!}{
\begin{tabular}{l| c| c}
\toprule
\textbf{Model Structure/Training Hyper-Parameters} &  \textbf{Name/Value} &  \textbf{More Information}  \\
\midrule
Base Model & \textit{LLaVA-ov-chat (7B)} &Model initialized from the base model\\
Vision Tower & \textit{SigLIP-SO400m} &\textit{Parameter size=}$397.75$M, \textit{Tokens per keyframe=$196$}  \\
Vision Projector & \textit{2-layers MLP+GeLU}&\textit{Parameter size}=$16.98$M \\
Motion Extractor& \textit{SlowFast-R50} &\textit{Parameter size=}$34.16$M, only use the fast path feature  \\
Motion Projector& \textit{2-layers MLP+GeLU}&\textit{Parameter size}=$13.77$M, the same with the structure of vision projector\\
LLM init. & \textit{Qwen-2 (7B)} &Decoder-only model, \textit{parameter size}=$7660.56$M \\
Keyframes Sampling Interval &1 second&/\\   
Keyframes Resolution         & $336\times 336$&/\\

Frames (for motion extraction) Resolution         & $224\times 224$&/\\
Batch Size (videos) & 8&\textit{Per device train batch size=1} \\
LR Max & 1e-5&/ \\
LR Schedule & cosine decay&/ \\
Warmup Epochs & $0.03$&/ \\
Weight Decay & $0$&/ \\
Gradient Accumulation Steps   & $1$&/ \\
Numerical Precision      & $\mathtt{bfloat16}$&/ \\
Epoch &$1$&/ \\
Optimizer & AdamW&/ \\
Activation Checkpointing &\checkmark&/ \\
Deepspeed Stage & $3$&/ \\
\bottomrule
\end{tabular}
}

\label{tab:hyperparamllava}

\end{table*}
\section{Detailed Information About Training }
\paragraph{Model Structure and Training Hyper-parameters.}
In Tab.~\ref{tab:hyperparamllava}, we provide a detailed overview of each model component's structural parameters and the hyperparameters used for training. \textbf{Specifically, we use the same hyperparameters in all $3$ training stages.}

\paragraph{Training and Evaluation Devices.}
The whole training process is conducted on $8$ NVIDIA H800-SXM5-80GB GPUs (requiring about $15$ hours in total).
All the evaluation experiments are conducted on $6$ NVIDIA A6000-48GB GPUs. 

\paragraph{Discussions on Specific Details.} As described in \ref{S/Amodels}, the training of our \textit{Assistant} model is based on the \textit{UGC-Scorer}, however bypassing the \textit{Stage-2} streaming video data. Additionally, we design distinct specialized \textit{Scorers} for typical UGC (offline) videos and streaming videos (featuring long-time stalling and rebuffering) instead of using a unified model. The reasons for this design are as follows.

We believe that extracting keyframes and motion features cannot accurately distinguish between long-time stalling/rebuffering (distinct from short-term stuttering/frame jitter) and the static-frame period during smooth playback. Precisely speaking, these two scenarios cannot be distinguished in an end-to-end manner without temporal annotations. Besides, viewers' quality judgments vary significantly depending on whether a long-time static period is identified as stalling. Therefore, combining streaming videos with long-term stalling (typically longer than $2$ seconds each time) with videos in other datasets free from such distortion type in a single training process may impair the model's ability to accurately differentiate between normal static playback and rebuffering states, causing performance drop in quality assessment and understanding. Thus, creating a specially designed model for evaluating the impact of long-term rebuffering/stalling on streaming video quality is a more suitable approach.

\section{Detailed Human Annotation Settings}
\label{sec:interface}

\subsection{Annotation Preparation} 
\begin{figure*}[h]
    \includegraphics[width=\textwidth]{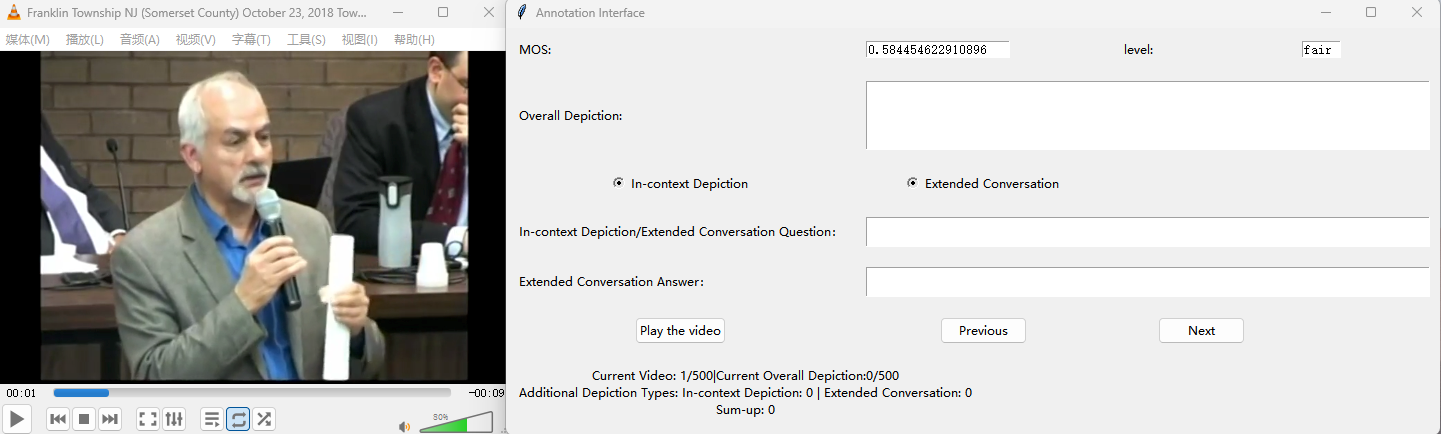}

    \caption{The interface allows subjects to provide feedback on different \textbf{pathways}. The normalized video MOS and the video quality level are presented to the subjects as references for quality attribute selection. }
    \label{fig:interface}

\end{figure*}

The device requires $1$-$2$ screens with a resolution of at least \textbf{1080p}. The screen can be a personal computer screen or monitor. To keep the size of the initially opened player consistent with the original resolution of the video, the  annotators need to use the \textbf{VLC player}. The interface screenshot is shown in Fig.~\ref{fig:interface}.

\label{sec:markbra}
\begin{table*}[h]\small
\caption{\textbf{The representative quality attributes examples.} The quality attributes in \textbf{bold} are those that must be considered in most videos. The quality attributes displayed in \textit{italic} font are the ones that can only be noticed in a small number of videos. The remaining attributes that are not marked should be considered depending on the content and quality level of the specific video.}

\centering
\begin{tabular}{p{6cm}|p{10cm}}
\hline
\textbf{Quality Attributes} & \textbf{Examples of Positive to Negative Representations} \\
\hline
\textbf{Sharpness} & Sharp, relatively clear, relatively fuzzy, very blurry \\ 
\hdashline
\textit{Focus}& In-focus, out-of-focus \\
\hdashline
Noise & Noiseless, a small amount of noise, severe noise densely distributed in the frame \\
\hdashline
Motion Blur & Clear-motion, blur-motion \\
\hdashline
\textbf{Flicker/Camera Shake} & Stable, small amplitude shake, shaky \\
\hdashline
\textbf{Exposure (subjective to light)} & Well-exposed, underexposure/overexposure \\
 \hdashline
\textbf{Compression Artifact} & Almost with no compression, with noticeable compression artifacts, severe compression blur, significant blockiness, serious loss of edge/texture details \\
\hdashline
\textbf{Fluency} & Smooth playback (fluent), (short-term) stutter/playback jitter \\
\hdashline
\textit{Contents} & Logically presented content, completely illogical \\
\hdashline
\textit{Composition} & Well-organized composition, poorly arranged frames, chaotic  \\
\hdashline
Color & Vibrant, natural, single, faded, unnatural\\
\hdashline
Light & Natural, soft, high contrast, uneven distribution, dark, underexposed, overexposed \\
\hdashline
Camera Trajectory & Coherent and stable, consistent, poor, shaky, lack consistency and coordination \\
\hdashline
\textit{Quality Switch} & Quality maintains at a consistent level, sharpness suddenly drops/rises at a certain moment, sharpness undergoes multiple violent changes within a certain time interval \\
\hline
\end{tabular}
\label{tab:qualityattributesexp}
\end{table*}

\begin{table*}[h]\small
\caption{\textbf{Degree descriptions and their criterion. }
To minimize the annotation bias, we provide the specific applicable range and criterion for each degree description level. (Given the unavoidable perceptual biases that exist with individual differences, the provided applicable ranges are intentionally designed to be flexible.) }
\vspace{-5pt}
    \centering
    \begin{tabular}{p{6cm}|p{10cm}}
        \hline
        \textbf{Degree} & \textbf{Criterion} \\
        \hline
        Very/Extremely Severe (used primarily for quality levels "Poor" and "Low") & The presence of such extremely severe distortions significantly degrades the viewer's quality of experience. The primary presence of such distortion leads to a very negative viewing experience, causing the overall experience to deteriorate significantly/important details in certain scenes are completely destroyed with such distortions. \\
        \hdashline
        Relatively/Quite Severe (used primarily for quality levels "Fair", "Poor" and "Low")  & The presence of such distortion moderately degrades the viewer’s overall viewing experience. This distortion is relatively prominent, annoying, and difficult to ignore. \\
        \hdashline
        Mild/Relatively Slight/Merely Noticeable (can be used in videos with all levels)  & The presence of such mild distortion has a relatively light but nontrivial impact on the viewer’s experience. \\
        \hdashline
        Good (used primarily for quality levels "High" and "Good") & This quality attribute is relatively good, and does not affect the viewing experience.\\
        \hdashline
        Excellent/Flawless (used only for the quality level "High") & This degree description is recommended when the quality attribute is outstanding and almost flawless while significantly enhancing the viewer’s experience. \\
        \hline
    \end{tabular}
    
    \label{tab:leveldispexp}
\end{table*}

\begin{table*}[h]\small
\caption{Temporal description examples.}
\vspace{-5pt}
    \centering
    \begin{tabular}{p{6cm}|p{10cm}}
        \hline
        \textbf{Description Type} & \textbf{Examples} \\
        \hline
        Overall Description & Appears throughout the entire playback / high frequency / multiple occurrences. \\
        \hdashline
        {In-context Temporal Description} &\textit{When}  Someone is doing something / \textit{When} event A occurs / \\ 
                                     & Camera switches \textit{from} Scene A \textit{to} Scene B  \\ 
\hdashline
{Direct Temporal Description} & At the very beginning of the video / At the $3$-rd second mark / \\ 
                                      &When the video is about to end / From $3$-$7$ seconds in the video \\ 
        \hline
    \end{tabular}
    \label{tab:temporaldispexp}
\end{table*}
\subsection{Overall Quality Depiction Settings}
\paragraph{The Specific Overall Depiction Settings.} Specific overall depiction settings are shown below:
\begin{itemize}
    \item The standard format of each quality attribute description is {\textbf{Quality Attribute}}$+${\underline{Degree}}$+$+{\textit{Temporal Description}} (the degree needs to be judged based on the perception of viewing and the quality level reference of the video). The quality attributes we recommend annotators to select are presented in Tab. \ref{tab:qualityattributesexp}. The example degree descriptions and their reference criterion are shown in \ref{tab:leveldispexp}.
    \item For temporal description, when a specific time point or period needs to be described, an \textit{in-context temporal description} has higher priority than a \textit{direct temporal description}. The examples of temporal descriptions are shown in Tab.~\ref{tab:temporaldispexp}.
    \item The overall quality depiction should clearly state the spatial area (the whole picture or a specific area) corresponding to each quality attribute under the temporal description. If a quality attribute exists in the whole frame (such as low sharpness caused by low resolution), such a location description can be omitted.
\end{itemize}

\paragraph{Quality-level-referenced Selective Description.}
To maximize the annotation value of medium-quality videos and reduce the difficulty for annotators while 
providing a more standardized criterion for selecting quality attributes in annotations to mitigate bias. We require the participants to identify quality attributes based on the reference quality level selectively. For medium-quality videos, annotators are instructed to identify more nuanced quality attributes corresponding to positive and negative aspects. The specific requirements are outlined as follows. In the examples below, the \textbf{bold}, \textit{italic}, and \underline{underlined} parts represent the \textbf{quality attributes}, \textit{temporal descriptions}, and \underline{degree descriptions}, respectively. Specifically, certain adjectives that reflect the positive or negative nature of a quality attribute can also be categorized as degree descriptions.

\label{sec:highlig}
\begin{itemize}
\item  \textbf{High (80-100)} Only describe the positive quality attributes that are the most impressive (Example: \underline{high} {\textbf{sharpness/resolution}} {\textit{throughout}}, the picture is {\textit{always}} \textbf{stable} and \underline{jitter-free}, the {\textbf{colors}} are \underline{very bright and natural} \textit{throughout}, the {\textbf{brightness}} is \underline{moderate} and the {\textbf{lighting}} is \underline{natural} \textit{throughout the playback}).

\item \textbf{Good (60-80)} This part of the videos gives good overall impressions, but most have noticeable flaws. The depiction process needs to consider both positive and negative quality attributes. The annotator should try to use a transitional tone, which is highlighted in \textcolor{red}{red} below (Example: The video has \underline{good} {\textbf{sharpness}} \textit{throughout the playback}, and the scene can be distinguished to a \underline{high} degree. \textcolor{red}{However}, there are {\textbf{flickers}} with  \underline{high frequency} and \underline{minor amplitude} during the video playback process, and the {\textbf{light}} is \underline{relatively dark and unnatural} \textit{when the camera switches to the woman in the middle}, which affects the viewer's viewing experience to a certain degree).
\item \textbf{Fair (40-60)} There are obvious flaws in this part of the video, and negative quality attributes (distortions) have become the main factor affecting the evaluation basis. The overall depiction only needs to describe the distortion present in the video (Example: the video has \underline{poor} {\textbf{clarity}} \textit{throughout} the playback process, there are \underline{high-frequency}, \underline{irregular}, and \underline{high-amplitude} {\textbf{camera shakes}}, and the {\textbf{color richness}} of \textit{in the second half of the video} is \underline{low}. These distortions seriously affect the viewer's viewing experience.) When the video has only a dominant distortion, a detailed description of this distortion can be added for extended depiction. (Example: There is \underline{relatively severe} {\textbf{compression blur}} \textit{throughout the video playback}, which makes the faces of the main characters in the video barely distinguishable, and the edge details are seriously lost).
\item \textbf{Poor (20-40)} This part of the video has more serious distortions. The annotation requirements are almost consistent with the \textbf{Fair level} (Example: The video has \underline{very poor} {\textbf{sharpness}} \textit{throughout the playback}. There are \underline{high-frequency}, \underline{irregular} and \underline{obvious} {\textbf{lens shakes}}, and the {\textbf{color richness}} \textit{during $3$-$7$ seconds} is very \underline{low}. These distortions seriously affect the viewer’s viewing experience).
\item{\textbf{Low (0-20)}} This part of the video has extremely low quality and is almost unwatchable. The requirements are the same as the previous two levels, and the degree description should be altered accordingly.
\end{itemize}

\subsection{In-context Depiction/Extended Conversation Settings}
As mentioned in \ref{annotation}, if the annotator chooses the \textbf{in-context depiction}, they are required to find one or more representative time points in the video/specific locations in the picture and describe their quality. If they need to describe a specific moment, the \textit{in-context temporal description} shown in Tab. \ref{tab:temporaldispexp} is preferable. 

If the distortions of the video are evenly distributed temporally and spatially, thus the in-context depiction is not suitable for this video, the annotators can choose to design an \textbf{extended conversation}.
This type of annotation has a wider range of content. Here are some examples:
\begin{itemize}
    \item Formulate a question and provide an answer about the cause of a particular type of distortion in the video.
    \item Design a question about the severity level of a specific distortion existing in the video, then respond with a relatively detailed answer. In addition to directly answering the severity, a brief supplementary explanation is recommended).
    \item Ask and answer about what improvement strategies could be applied during reshooting to achieve a higher-quality video with the same content.
    \item Consider what post-processing techniques might be used to enhance the quality of the video with distortions, ensuring the same content but improved quality.
    \item Create a question and response about a specific quality attribute of the video, such as light, color temperature, color accuracy, edge detail, overall sharpness, etc.
    \item Design a question and response focusing on the quality of a particular moment, period, or certain spatial location within the video (similar to the in-context depiction rephrased in a conversational format).
\end{itemize}

\subsection{Special Designs for AIGC-video Annotations}
Since AIGC video quality evaluation focuses more on visual realism, factual consistency, and motion coherence—rather than solely on low-level visual quality—our approach to creating overall quality depictions for the $998$ AIGC videos in the dataset differs from the previous requirements. Specifically, we instruct annotators to provide detailed descriptions across $5$ aspects provided in the original dataset \cite{he2024videoscore}: \textit{visual quality}, \textit{temporal alignment}, \textit{dynamic degree}, \textit{text-to-video alignment}, and \textit{factual consistency}. We use the quality levels (ranging from $1$ to $4$) annotated in the dataset as references for overall depictions. Additionally, given the fact that the length of these videos is generally very short (typically less than $5$ seconds with a low frame rate), the \textit{temporal description} is not necessary. Additionally, we bypass the in-context depiction / extended conversation annotation step. The \textit{GPT} rewriting process is also solely based on the $5$ dimensions in the overall depiction.
\subsection{Overview of Subjective Experiment Setup}
The subjective annotation experiment involves $33$ participants, all well-educated undergraduate or graduate students with relevant professional backgrounds. Among them, $16$ participants are experts in related fields. The selection of participants ensures the quality of the human-annotated data. The entire experiment lasts one month, during which all participants are required to view and annotate videos in environments meeting the requirements of formal subjective experiments. The number of videos annotated by each participant ranges from $100$ to $1,000$. To prevent fatigue, participants are instructed to evenly distribute their daily annotation tasks, with no less than $15$ videos and no more than $30$ videos per day.

\section{Prompts Design}
\subsection{Prompts design for GPT-extension}
We extend and refine the human annotations by sending prompts to \textit{GPT}. The following are the prompts we use in detail, where \textit{GPT} is referred to as ``you".

\paragraph{Quality-centric Q\&A Pairs Rewriting Prompt.}
You will receive a detailed depiction of the quality of a video, which may include evaluations and comments on various attributes of the video's quality. Based on this depiction, create exactly $3$ questions about specific video quality attributes (sharpness, fluency, flicker, etc.) and provide the corresponding answers. The questions and answers can be in the form of yes/no choices (e.g., Question: Is the sharpness of the video high throughout the playback? A. Yes B. No Answer: A.), multi-choices (single answer) questions (e.g., Question: How is the degree of the sharpness of the video throughout the playback? A. High B. Relatively good C. Relatively poor D. Extremely low. Answer: B.) or open-ended responses (e.g., Question: How would you rate the sharpness of the video throughout playback? Answer: Average/Excellent/Poor). Please ensure that the questions and answers are entirely based on the video quality depiction without including any information beyond it. Below is an example:

\begin{itemize}
    \item \textbf{Overall Quality Depiction:} The video playback is smooth and fluent. In the first half, the picture is stable without flicker when the video is indoors. However, when the scene switches to the outdoor playground, the flicker becomes noticeable, and the sharpness and overall brightness of the image decrease significantly. (Additionally, there is a certain level of compression blur throughout the video.)
\end{itemize}

\begin{itemize}
    \item \textbf{Question 1}
 
    Question: Is the video playback smooth and fluent? A. Yes B. No.

    Answer: A. Yes

    \item \textbf{Question 2}

    Question: How is the stability in the first half of the video when indoors? A. Very poor, with severe flickers. B. The stability is good, with no flicker. C. There are moderate but noticeable camera shakes. D. It is relatively good; however, the video still has slight flickers.

    Answer: B. The stability is good, with no flicker.

    \item \textbf{Question 3}

    Question: How is the sharpness throughout the video playback?

    Answer: The sharpness is relatively poor due to the compression blur throughout the video.
\end{itemize}

Please note that questions like ``How is the brightness of the video?' are not suitable as there is no absolute description of brightness quality in the description, only a mention of its decrease. Additionally,  you will not see the video; only the textual depiction will be provided. Therefore, all questions and answers must strictly adhere to the depiction, and no information beyond what is described should be included. Please format the output as follows: Question 1: Question: Answer: Question 2: Question: Answer: Question 3: Question: Answer:. You need to simulate as if you have seen the video; therefore, all questions and answers should not include any traces of the provided text depiction.

\paragraph{Temporal-centric Q\&A Pairs Rewriting Prompt.}
You will receive a detailed depiction of the quality of a video, which may include evaluations and comments on various attributes of the video quality. Based on this depiction, create a question related to the temporal aspect of the video and provide the corresponding answer. The question types include: 

\begin{enumerate}
    \item \textbf{Temporal analysis}, which is about and answers the relationship between the quality of two different time points or segments within the video. 

    \textbf{Example:} Depiction: The video playback is smooth and fluent. In the first half, when the video is indoors, the picture is stable without flicker. However, when the scene switches to the playground, the flicker becomes noticeable, and both the sharpness and overall brightness of the image decrease significantly. Question: Which part has higher overall brightness, the first half indoors or the second half when switched to the playground? Answer: The first half when indoors.

    \item \textbf{Temporal retrieval}, which is about when a quality event occurs in the video or how frequently it appears.
    
    \textbf{Example 1:} Depiction: Severe overexposure occurred before the scene transition at the beginning of the video playback. Question: At what point does severe overexposure occur in this video? Answer: Before the scene transition at the beginning of the video. 
    
    \textbf{Example 2:} Depiction: High-frequency flickering occurs throughout the video playback. Question: Do flickers occur during the video playback, and how frequently? Answer: Yes, there are flickers that occur with high frequency.
\end{enumerate}

Please note that you will not see the video itself but only the text depiction provided, so all questions and answers must strictly adhere to the depiction. If the description does not include relevant information, such information should not appear in the questions or answers you design. You may design questions through single-answer choices or open-ended responses. Please format the output as follows: Question: Answer:. The video quality depiction is: \textbf{[overall depiction]}. You need to simulate as if you have seen the video; therefore, all questions and answers should not include any traces of the provided text depiction.

\paragraph{Extended Conversation Rewriting Prompt (From Overall Depiction).}
You will receive a detailed depiction of the quality of a video, which may include evaluations and comments on various attributes of the video quality. Based on this depiction, create only $1$ extended question and provide an answer. The question types include but are not limited to:

\begin{enumerate}
    \item Requesting an inference about the reason for a particular quality attribute (positive or negative) observed in the video.

    \textbf{Example:} Depiction: The video captures the person’s face clearly, and the person remains close to the camera and directly faces it throughout. Question: Why is the person's face captured clearly in this video? Answer: The person’s face is close to the camera and directly to it.

    \item Requesting a specific post-processing method that can improve the quality of a distorted video based on the negative attributes described.

    \textbf{Example:} Depiction: The video has low sharpness throughout playback, with severe compression blur and block artifacts, and exhibits persistent, irregular flicker. Question: What post-processing methods could be used to improve the quality of this video and reduce distortion effectively? Answer: Video super-resolution can be applied to reduce the impact of compression blur and block artifacts, and de-flicker operations can reduce flicker effects.

    \item Requesting a specific strategy for re-shooting the video to achieve a higher quality version of the same content based on the negative aspects described. 

    \textbf{Example:} Depiction: The video has low brightness and faded colors. Question: What adjustments could be made to re-shoot the video to achieve higher quality? Answer: Reshooting the video on a sunnier day or increasing the overall brightness of the surrounding shooting environment would improve quality.
\end{enumerate}

Please note that you will not see the video; only the textual depiction will be provided. Therefore, all questions and answers must strictly adhere to the content of the depiction. If the depiction explicitly states the cause of a quality event, questions of type $1$ must strictly follow the reason given in the depiction. If no explicit cause is stated in the depiction, do not design type $1$ questions. Please format the output as follows: Question: Answer:. The video quality depiction is: \textbf{[overall depiction]}. You need to simulate as if you have seen the video; therefore, all questions and answers should not include any traces of the provided text depiction.

\paragraph{Incontext Depiction Q\&A Pairs Rewriting Prompt.}
You will receive one brief in-context quality depiction, which primarily describes a positive or negative quality phenomenon occurring in a specific moment or period of a video or space within the frames. Based on this depiction, please rewrite only $1$ question. Requirements: If the in-context quality description targets a specific moment or period in the video or a particular object/part of the frame space (or a combination of both), the rewritten question must focus on that specific time or spatial part/object stated in the depiction. 

\textbf{Example 1:} Depiction: Out-of-focus blur occurs when the person starts running in the video. Question: At what moment does defocus blur occur in the video? Please describe this moment precisely. Answer: When the person starts running in the video.) 

\textbf{Example 2:} Depiction: There is local overexposure on the full-body of the performers in the center and on the right. Question: Where in the frame does overexposure occur? Answer: On the full-body of the performers in the center and on the right side.) 

Please note that you will not see the video but only the textual depiction provided. Therefore, all questions and answers must strictly adhere to the depiction, and no information beyond what is described should be included in the questions and answers. Please format the output as follows: Question: Answer:. You need to simulate as if you have seen the video; therefore, all questions and answers should not include any traces of the provided text depiction.

\subsection{System Prompts Design}
As mentioned in \ref{setup}, a specific system prompt is added before the instruction Q\&A pairs to facilitate the LMM's understanding of the exact meaning of the question and its corresponding answer.

\paragraph{System Prompt in Training State.}
Across the $3$ training stages, the design of all system prompts remains entirely consistent.

\noindent\textbf{\textit{System Prompt:}}\textit{ Now you will receive one video. This video is \textbf{\textit{[length]}} seconds long, and you will see a sequence of keyframes generated by uniformly sampling $1$ frame per second from the video. The keyframe sequence follows the original order of the video. After uniform sampling, there are a total of \textbf{\textit{[length]}} images: \textbf{\textit{[image]}}. In addition, you will also obtain motion features extracted from all \textbf{\textit{[num frames]}} frames of the entire video: \textbf{\textit{[motion]}}. The temporal motion features also follow the original frame order of the video. Please watch this video carefully, and then answer the following question.} 

\paragraph{System Prompt for Quality Scoring.} When evaluating quality scoring tasks, we choose a relatively simple system prompts design to guarantee that the evaluation process is end-to-end (without any additional temporal information that may need manual extraction, like length, frame rate, and stalling information). The system prompt design is shown below.

\noindent\textbf{\textit{System Prompt:}} \textit{The key frames of this video are: \textbf{\textit{[image]}}. And the motion feature of the video is \textbf{\textit{[motion]}}. Please watch this video carefully, and then answer the following question.}

\paragraph{System Prompt for Quality Understanding.} When evaluating quality understanding tasks, we provide some temporal information to the system prompt since answering some questions in the benchmark precisely needs access to that temporal information.

\noindent\textbf{\textit{System Prompt:}} \textit{You will receive \textbf{\textit{[length]}} distinct frames that have been uniformly sampled in each second from a video frames sequence, arranged in the same temporal order as they appear in the video. In addition, you will also obtain motion features extracted from all \textbf{\textit{[num frames]}} frames of the video.  Please analyze these images and motion features and answer the question based on your observations. The video frames: \textbf{\textit{[image]}}, the motion features \textbf{\textit{[motion]}}}.

\section{Quality Scoring Method}
We use the following method to get the predicted quality score during the evaluation of the quality scoring tasks:
\begin{equation}
\mathcal{Q}=\sum_{i=1}^5\omega_i\frac{e^{\mathcal{P}_{\textit{quality\_levels}[i]}}}{\sum_{i = 1}^5e^ {\mathcal{P}_{\textit{quality\_levels}[i]}}},
\end{equation}
where \textit{quality\_levels} denotes the list of transformed quality levels: \textit{[High, good, fair, poor, low]} and $\mathcal{P}$ represents the model's output \textbf{logit} of each quality level. Specifically, we first extract the vector (with its dimension matching the tokenizer vocabulary size) at the model's output sequence corresponding to the quality-level description word (the $-3$ index in our model). We then take the \textbf{logits} at the indexes in this vector associated with each quality level in the tokenizer's vocabulary (indexes $1550$, $1661$, $6624$, $7852$, and $3347$ in our model).
Then, we apply softmax normalization to these $5$ logits. $\omega$
denotes the weights assigned to the normalized probabilities of these $5$ quality levels ($[1,0.75,0.5,0.25,0]$). The weighted sum of these probabilities yields the predicted quality score $\mathcal{Q}$, ranging between $[0,1]$.
\begin{figure*}[h]
    \centering
    \includegraphics[width=0.7\textwidth]{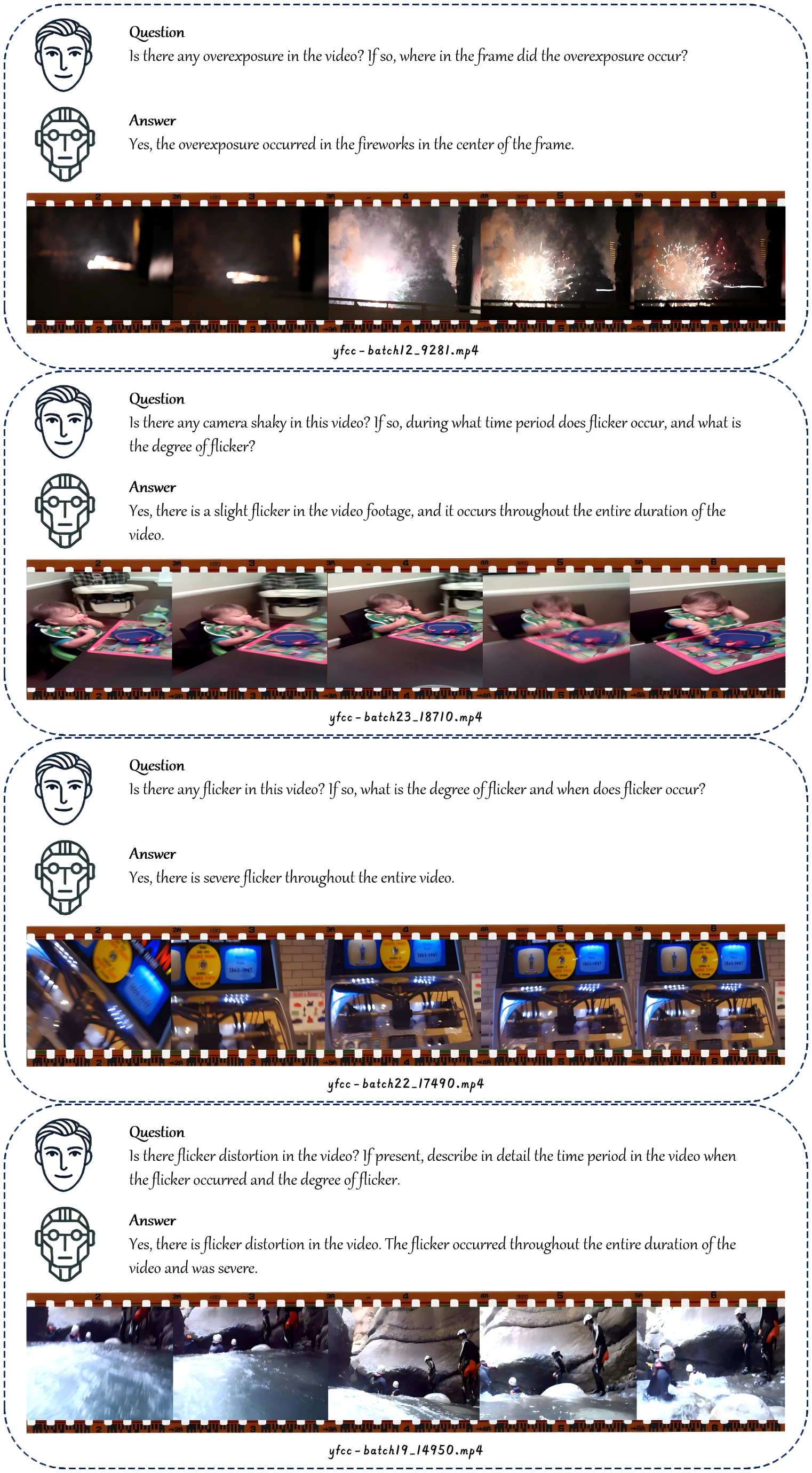}
    \vspace{-10pt}
    \caption{\textbf{Qualitative Examples (I)}}
    \label{fig:appendix1}
    \vspace{-10pt}
\end{figure*}

\begin{figure*}[h]
    \centering
    \includegraphics[width=0.7\textwidth]{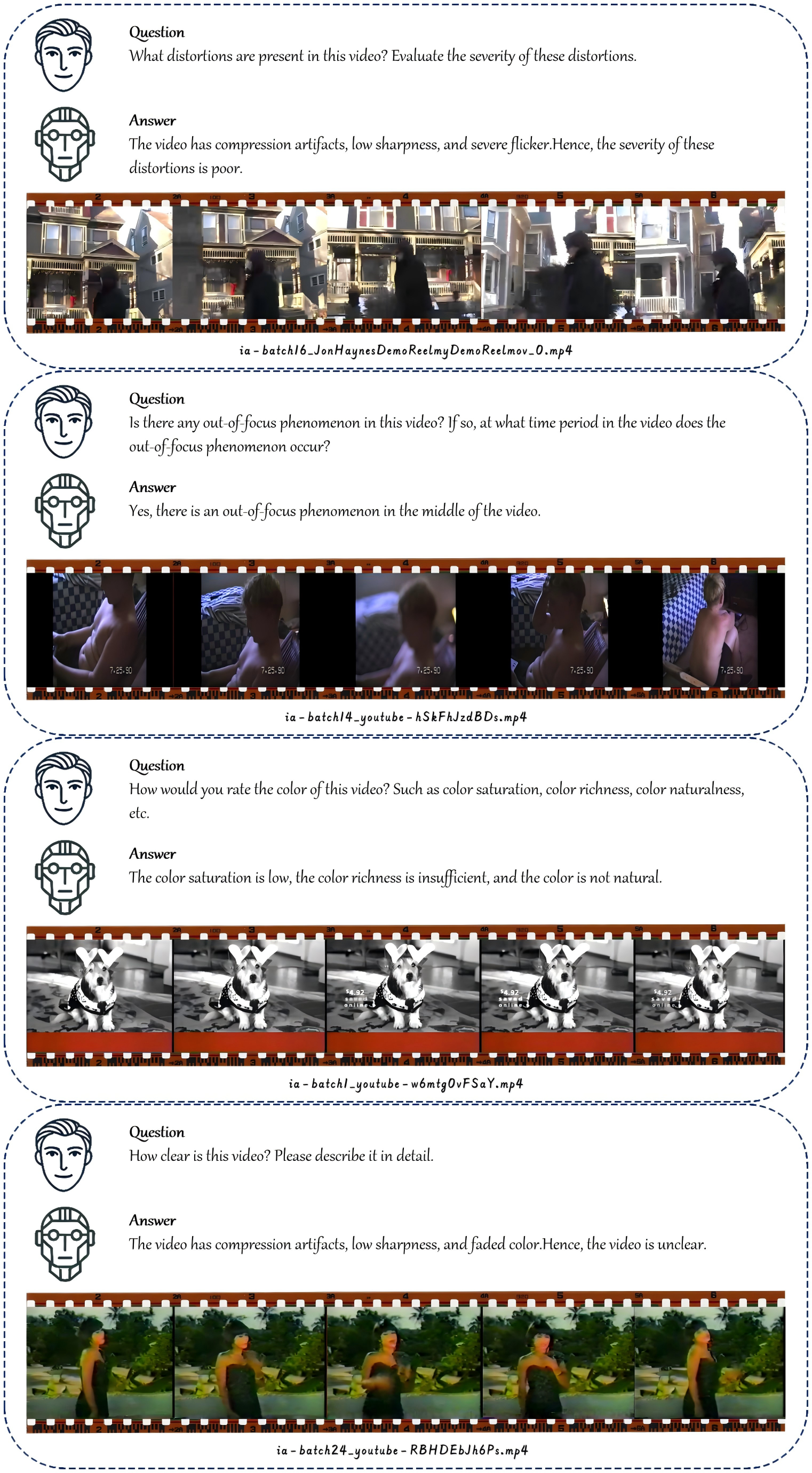}
    \vspace{-10pt}
    \caption{\textbf{Qualitative Examples (II)}}
    \label{fig:appendix2}
    \vspace{-10pt}
\end{figure*}
\begin{figure*}[tb]
    
    \centering
    \includegraphics[width=0.95\textwidth]{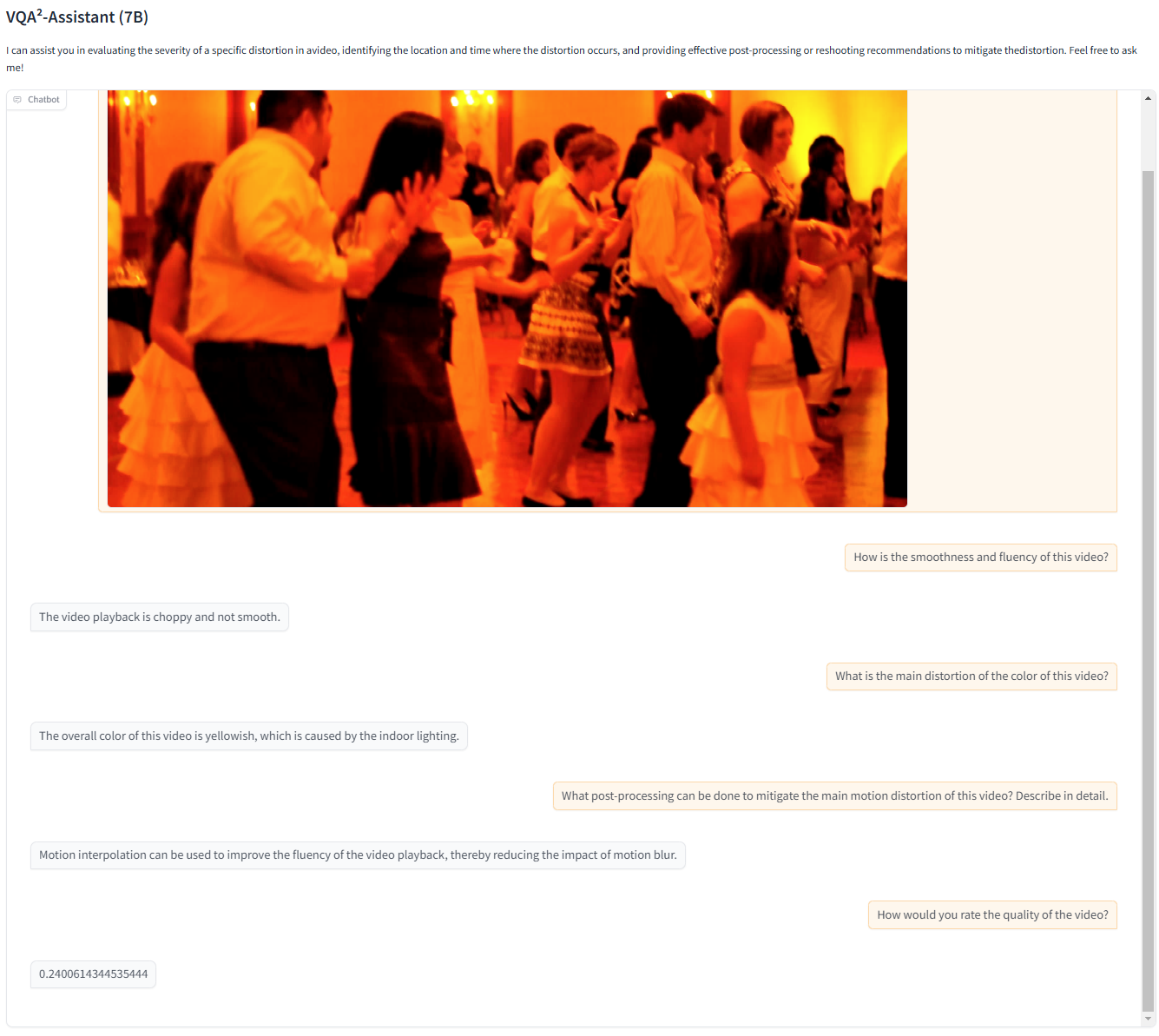}
   
    \caption{A video with extremely high color temperature, very low sharpness, and frequent, severe stuttering. The case demonstrates that the \textit{Assistant} model can accurately identify videos with low fluency, perceive the level of color temperature, give reasonable suggestions for quality refinement, and provide a relatively accurate scoring for low-quality videos.}
    \label{fig:case1}

\end{figure*}
\begin{figure*}[t]
 
    \centering
    \includegraphics[width=0.95\textwidth]{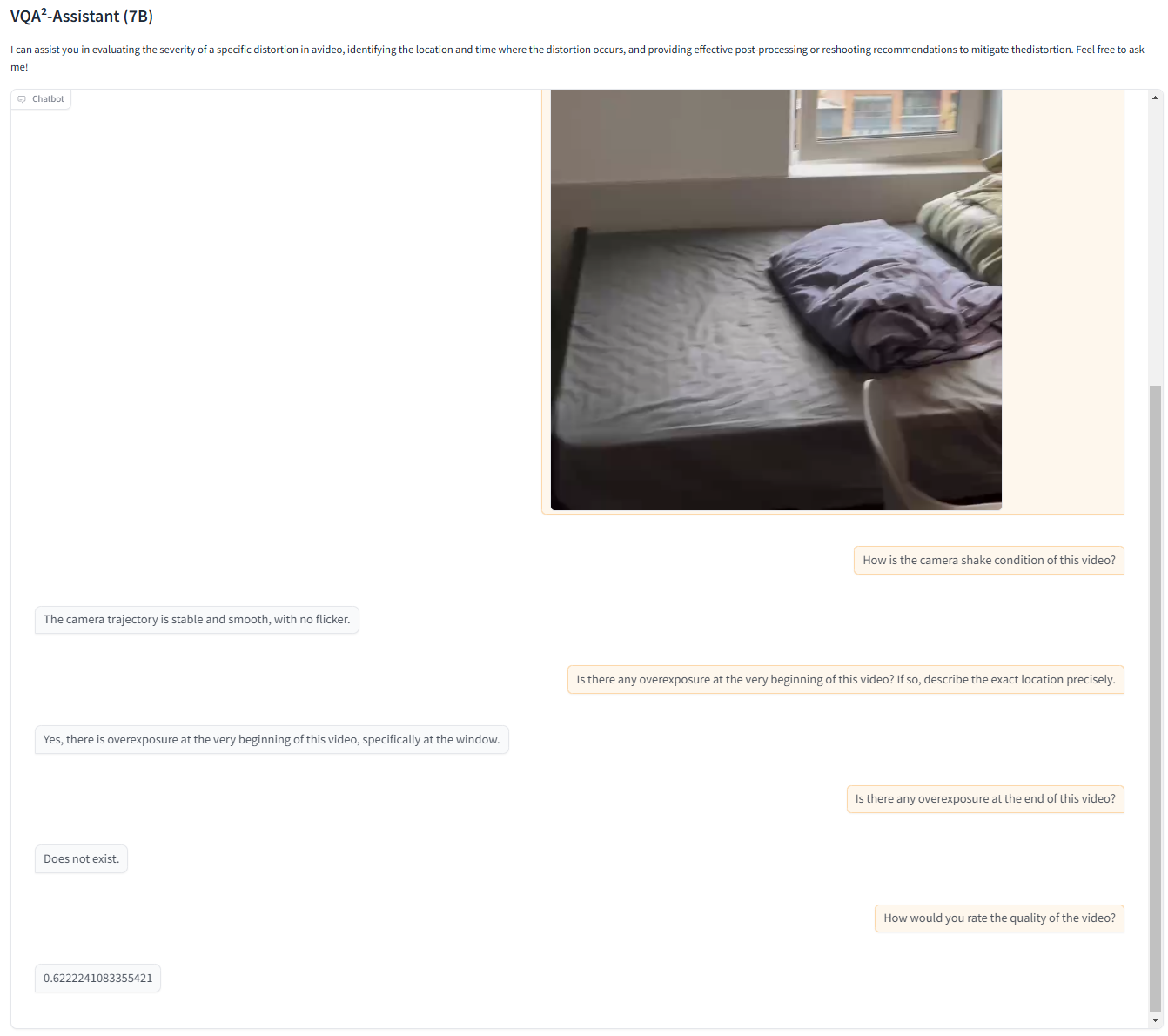}
   
    \caption{A video shot on a mobile phone with a stable camera trajectory. The video undergoes some compression during file transfer, slightly reducing its sharpness. At the beginning, the video shows noticeable overexposure on the window, which does not persist throughout, resulting in overall medium quality. This example demonstrates that the \textit{Assistant} model can accurately detect camera shake, perform precise spatial-temporal retrieval or analysis of distortions, and provide a reasonably accurate rating for a medium-quality video.}
    \label{fig:case2}

\end{figure*}
\begin{figure*}[t]
 
    \centering
    \includegraphics[width=\textwidth]{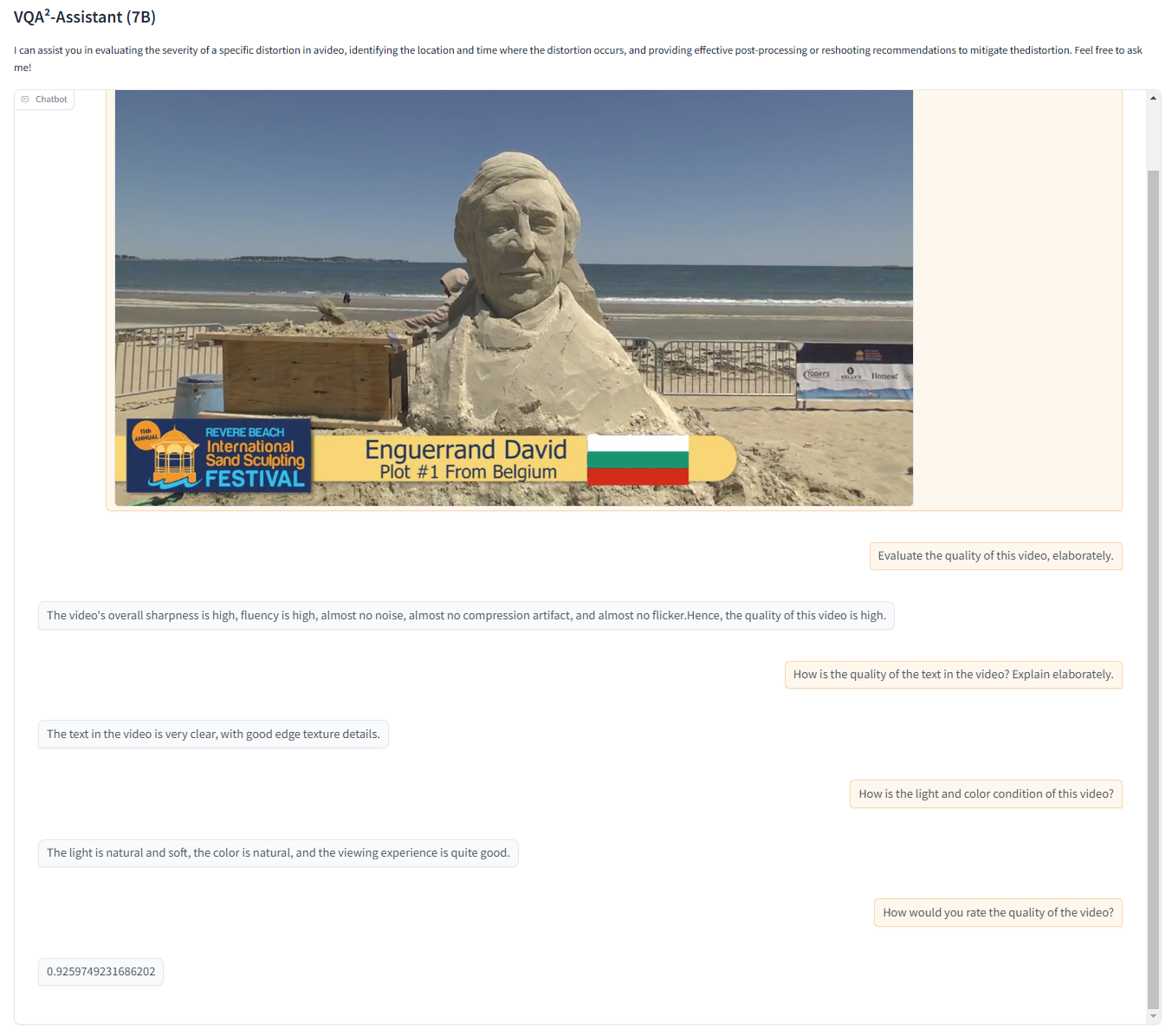}
   
    \caption{A high-quality video with overall high sharpness, natural lighting and colors, and sharp, high-resolution text with clear edges. This case demonstrates that the \textit{Assistant} model can accurately assess multiple quality attributes in a high-quality video and provide a reasonable rating for high-quality videos.}
    \label{fig:case3}

\end{figure*}
\begin{figure*}[t]
 
    \centering
    \includegraphics[width=\textwidth]{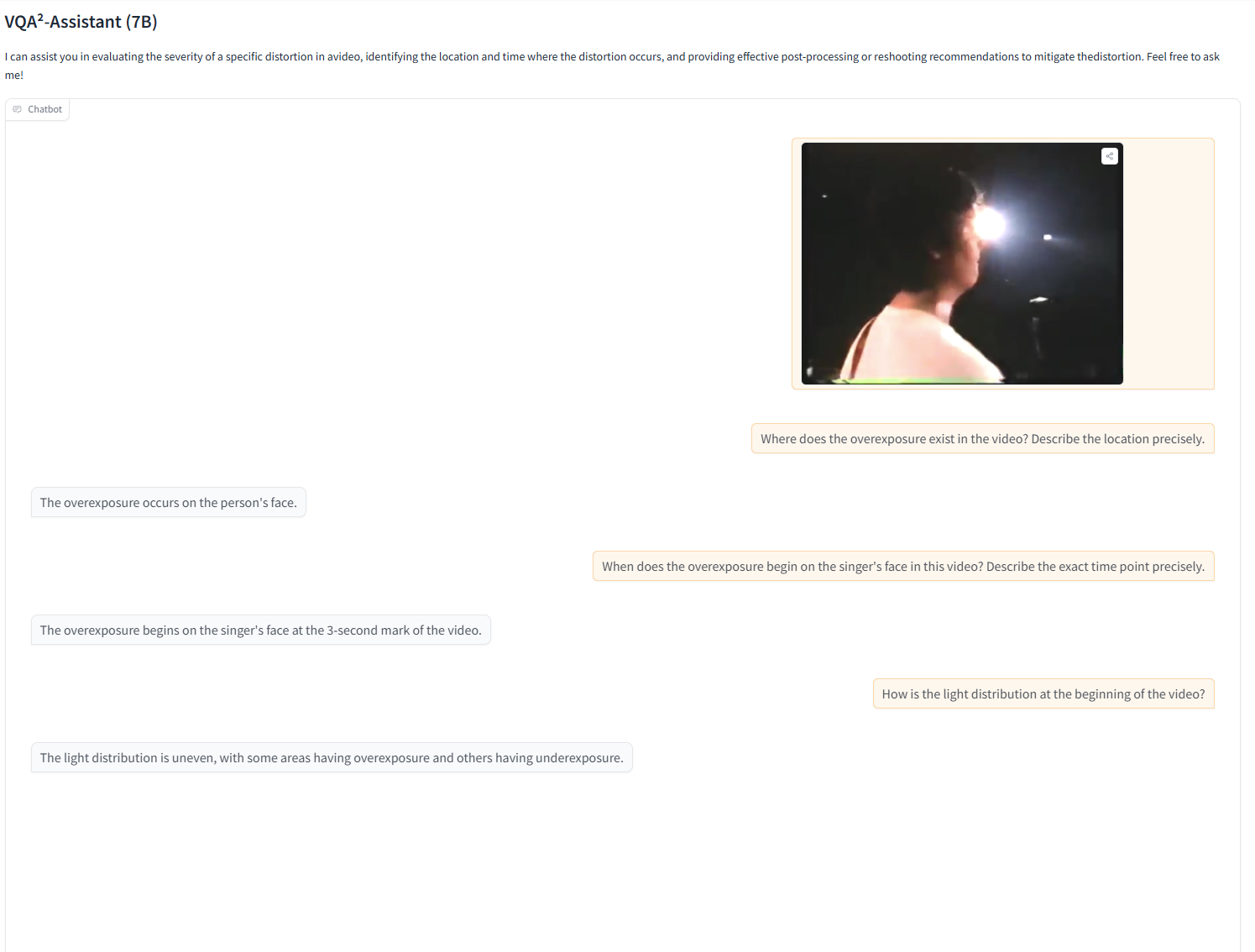}
   
    \caption{A low-quality video with uneven light throughout, with some areas overexposed and others completely dark. Noticeable overexposure begins on the singer's face and background light around the 3-second mark. This case illustrates that the \textit{Assistant} model has relatively strong local in-context distortion analysis capabilities and can accurately perform temporal distortion retrieval down to specific seconds.}
    \label{fig:case4}

\end{figure*}
\section{Qualitative Analyses}
We conduct a detailed qualitative test on this model to demonstrate the \textit{Assistant}'s strong quality understanding and question-answering capabilities. We present $8$ videos selected from the LSVQ dataset featuring unique motion or spatial distortions (all excluded from \textit{Stage-3} training videos). The questions designed include analyzing distortion types and overall video quality, providing qualitative assessments, temporal/spatial retrieval of distortion locations, and describing specific quality attributes in detail. Screenshots of the test video frames, along with the test results, are shown below (in Fig. \ref{fig:appendix1} and \ref{fig:appendix2}).


\section{Applications}
We created a demo based on \textbf{\textit{Gradio}} to test the application of the \textit{Assistant} model in video quality question-answering. We provide four specific examples in Fig. \ref{fig:case1}, \ref{fig:case2}, \ref{fig:case3}, and \ref{fig:case4}) .

\section{Limitations}
Due to time constraints, we can not conduct extensive experiments to obtain a larger amount of distortion recognition pretraining data. We utilize the data from existing studies and some synthetically generated data. Therefore, the \textit{Stage-1} can be regarded as a "task-specific pretraining for video quality assessment" on small-scale data rather than the large-scale pretraining typically used to construct foundation models in comprehensive LMMs. 

Additionally, the strict and detailed annotation requirements, combined with the manual review and revision of all human annotations to eliminate unqualified entries, significantly limit the time available for scaling up the \textit{Stage-3} instruction set, as these processes are essential to ensuring high annotation quality. Thus, the scaling-up of model size is also affected by this. 

Despite employing various methods, such as \textit{GPT}-based extension and refinement and quality-level-referenced attribute selection to minimize annotation biases, the individual perception biases are impossible to eliminate entirely. This limitation slightly impacts the diversity of model outputs and the accuracy of responses to specific questions.

Furthermore, since the series models (especially the \textit{Assistant}) are specifically designed for low-level video quality assessment tasks, especially for the UGC videos, their performance and generalization ability might face a little challenge on other general-purpose tasks/benchmarks and video content types.

These limitations highlight the key points for further exploration.

\section{Acknowledgements}
In the subjective annotation experiments, all participants are clearly informed in advance about the workload. Participation in the experiment is entirely voluntary without any enforcement. Throughout the experiment, no participants report experiencing fatigue or discomfort. Upon completion of the experiment, we provide all participants with appropriate labor wages. We would like to express our heartfelt gratitude to all participants for their contributions to the subjective experiments. 
\section{License}
All the outputs of our work, including the complete data of the \textbf{\textit{VQA\textsuperscript{2} Instruction Dataset}} as well as the training and evaluation code and fine-tuned weights of the \textbf{\textit{VQA\textsuperscript{2} Series Models}}, will be fully open-sourced and made freely available to the research community for further studies.
\end{document}